\documentclass{article}

\usepackage{microtype}
\usepackage{graphicx}
\usepackage{subcaption}
\usepackage{booktabs}
\usepackage{hyperref}
\usepackage{amsmath}
\usepackage{amssymb}
\usepackage{mathtools}
\usepackage{amsthm}
\usepackage{framed} 
\usepackage{multirow}
\usepackage{enumitem}
\usepackage{caption}
\usepackage{float} 
\usepackage{etoolbox}
\usepackage{xspace}
\usepackage{accents}
\usepackage{makecell}
\usepackage{bbding}
\usepackage{multicol}
\usepackage[normalem]{ulem}
\usepackage[capitalize,noabbrev]{cleveref}
\usepackage[textsize=tiny]{todonotes}
\usepackage{xcolor}
\usepackage{colortbl}

\usepackage[preprint]{icml2026}

\theoremstyle{plain}

\theoremstyle{definition}

\theoremstyle{remark}

\theoremstyle{problem}

\newcommand{\eg}{\emph{e.g.},\xspace}

\newcommand{\model}{MADI}

\newcommand\figref[1]{Figure~\ref{#1}}

\newcommand\tabref[1]{Table~\ref{#1}}
\newcommand\secref[1]{Section~\ref{#1}}
\newcommand\equref[1]{Equation~(\ref{#1})}
\newcommand\appref[1]{Appendix~\ref{#1}}

\newcommand{\eat}[1]{}

\icmltitlerunning{Aligned and Disentangled Multi-modal Learning for Time Series Understanding and Reasoning}

\begin{document}

\twocolumn[
  \icmltitle{From Consistency to Complementarity: Aligned and Disentangled Multi-modal Learning for Time Series Understanding and Reasoning}



  \icmlsetsymbol{equal}{*}

  \begin{icmlauthorlist}
    \icmlauthor{Hang Ni}{hkust}
    \icmlauthor{Weijia Zhang}{hkust}
    \icmlauthor{Fei Wang}{ict,ucas}
    \icmlauthor{Zezhi Shao}{ict}
    \icmlauthor{Hao Liu}{hkust}
  \end{icmlauthorlist}

  \icmlaffiliation{hkust}{The Hong Kong University of Science and Technology (Guangzhou)}
  \icmlaffiliation{ict}{Institute of Computing Technology, Chinese Academy of Sciences}
  \icmlaffiliation{ucas}{University of Chinese Academy of Sciences}

  \icmlcorrespondingauthor{Hao Liu}{liuh@ust.hk}

  \icmlkeywords{Machine Learning, ICML}

  \vskip 0.3in
]

\printAffiliationsAndNotice{}

\begin{abstract}
Advances in multi-modal large language models (MLLMs) have inspired time series understanding and reasoning tasks, that enable natural language querying over time series, producing textual analyses of complex temporal dynamics. 
Recent attempts hybridize numerical time series with their visualized plots, facilitating precise value reasoning and visual structure comprehension for comprehensive time series understanding of MLLMs.
However, effective numerical-visual modality integration remains challenging due to fine-grained temporal misalignment across modalities and severe entanglement between shared and modality-specific semantics, which hinder localized interpretation and complementary reasoning.
To address these issues, we propose \textbf{\model}, a \underline{\textbf{m}}ulti-modal LLM enhanced with fine-grained \underline{\textbf{a}}lignment and \underline{\textbf{d}}isentangled \underline{\textbf{i}}nteraction, featuring (1) Patch-level Alignment, which enforces physically grounded fine-grained correspondence across heterogeneous modalities, (2) Discrete Disentangled Interaction, which separates modality-common semantics into compact discrete latents and adaptively synergizes the purified modality-unique information, and (3) Critical-token Highlighting, which emphasizes informative, query-relevant signals for robust reasoning.
Experiments on synthetic and real-world benchmarks show that \model\xspace consistently outperforms general-purpose LLMs and time-series-specialized MLLMs.
\end{abstract}

\section{Introduction}
\label{sec:intro}

Time series understanding and reasoning (TSUR) tasks aim to interpret natural language queries alongside uni- or multi-variate time series inputs, producing textual explanations or analyses of temporal dynamics~\cite{chatts2025}. 
Unlike classical time series tasks (\eg forecasting, classification, anomaly detection)~\cite{wang2024deep} which operate on uni-modal numerical inputs and yield task-specific outputs with limited flexibility and interpretability, TSUR supports interpretable, language-driven interactions between time series data and users via open-ended natural language dialogue.
Such capabilities are crucial for human decision-making in real-world scenarios, including healthcare~\cite{gem2025}, finance~\cite{chen2025mtbench}, and industrial maintenance~\cite{itformer2025}.

\begin{figure}[t]
    \centering
    \includegraphics[width=\linewidth]{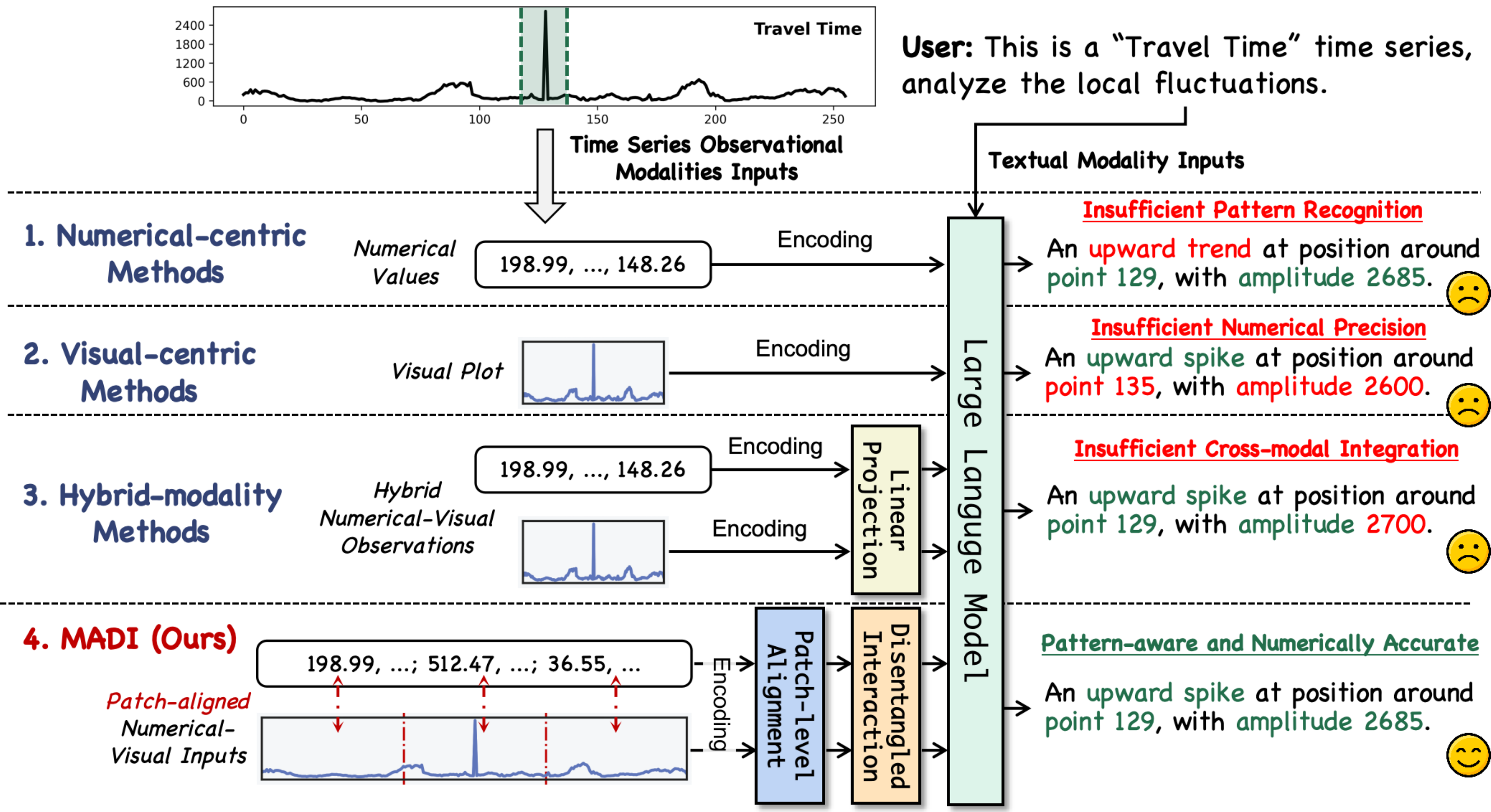}
    \vspace{-15pt}
    \caption{Comparison of multi-modal architectures.}
    \vspace{-20pt}
    \label{fig:motivation}
\end{figure}

With the rapid advancement of large language models (LLMs), TSUR tasks have become dominated by LLM-based paradigms, enabling flexible user queries and textual outputs. Beyond linguistic inputs, existing works primarily focus on \emph{how to formulate time series into effective observational modalities} that can be jointly processed with language by LLMs.
\emph{Numerical-centric methods} directly feed raw time series values into LLMs, either by encoding them as plain text via the LLM tokenizer and embeddings~\cite{vltime2025,merrill2024language}, or by introducing specialized time series encoders as an additional modality~\cite{chattime2025,itformer2025,chatts2025}.
While preserving precision, studies indicate that LLMs struggle to recognize high-level temporal structures like trends and periodicity~\cite{vltime2025}, reflecting a fundamental numerical-language modality gap induced by text-only LLM pre-training~\cite{tan2024language}.
In contrast, \emph{visual-centric methods} convert time series into visual plots and leverage the strong visual processing abilities of multi-modal LLMs (MLLMs)~\cite{vltime2025,merrill2024language,timemaster2025}.
Although effective at capturing high-level temporal structures, these approaches often sacrifice fine-grained numerical details that are critical for precise temporal analysis~\cite{visualanomaly2025,zhao2025images}.
Recent hybrid approaches like GEM~\cite{gem2025} attempt to combine numerical and visual modalities within MLLMs via token concatenation for comprehensive time series understanding.
Despite these advances, effectively integrating these two observational modalities for MLLMs reveals two critical challenges:

\textbf{(1) Fine-grained Cross-modal Alignment.}
Numerical values and their visual renderings reflect the same underlying physical signal, yet they are often encoded independently, leading to misaligned representations that hinder cross-modal reasoning.
Moreover, vision-language pre-training gaps of MLLMs amplify this misalignment, preventing models from achieving the interpretive accuracy with numerical data that they attain in vision.
Existing methods either rely on implicit alignment via linear projections~\cite{gem2025}, which can not enforce explicit physical correspondence, or perform series-level contrastive alignment~\cite{timesclip2025}, which overlooks fine-grained temporal variations (\eg fluctuations, noise) and can induce localized hallucinations.
Hence, a key challenge is to establish physically grounded, fine-grained alignment between numerical and visual modalities, providing a consistent semantic reference for accurate cross-modal understanding.

\textbf{(2) Disentangled Cross-modal Interaction.}
Although alignment ensures fundamental cross-modal consistency, it is not sufficient for effective time series understanding without fully leveraging complementary strengths of visual patterns and numerical details.
To integrate these unique features across modalities, existing approaches employ various multi-modal fusion and interaction strategies, such as feature concatenation~\cite{gem2025}, additive fusion~\cite{timevlm2025} and cross-attention~\cite{timecma2025}.
However, numerical and visual modalities describe the same underlying signals, resulting in substantial information overlap entangled with the modality-unique semantics.
Directly operating on full-modal representations often allows redundant, modality-shared content to dominate, diluting the unique contributions of each modality and limiting cross-modal synergy.
This motivates the second challenge: how to disentangle shared and modality-specific semantics after alignment, and integrate complementary signals to achieve synergistic reasoning, rather than mere feature aggregation.

To address these challenges, we propose \textbf{\model}, a \underline{\textbf{m}}ulti-modal LLM (MLLM) for TSUR tasks, enhanced with fine-grained \underline{\textbf{a}}lignment and \underline{\textbf{d}}isentangled \underline{\textbf{i}}nteraction across heterogeneous modalities.
The distinctions between \model\xspace and prior methods are highlighted in \figref{fig:motivation}.
We first introduce a \emph{Patch-level Alignment (PA)} module that performs fine-grained contrastive alignment between numerical time series and their visualized line plots at a patch granularity, ensuring physically grounded local correspondence for accurate cross-modal reasoning.
To enhance MLLM comprehension of numerical inputs, we further incorporate patch-wise textual captions derived from the time series, aligning numerical values with their textual descriptions.
Together, these alignments ground numerical representations in the MLLM-compatible vision–language semantic space, facilitating robust multi-modal reasoning.
Building on the aligned representations, we devise a \emph{Discrete Disentangled Interaction (DDI)} module to exploit the complementary benefits of numerical precision and visual abstraction.
To separate modality-unique semantics from modality-common redundancies, DDI introduces a compact discrete latent space via hierarchical vector quantization, capturing multi-scale, compact modality-common features to enhance disentanglement purity.
The unique components are then integrated through cross-attention, enabling synergistic interaction while preventing interference from redundant shared signals.
Finally, to enhance generalization, we introduce a lightweight \emph{Critical-token Highlighting (CTH)} module that identifies informative, question-relevant tokens across modalities, and prepends them to the token sequences.
By incorporating these modules into a pre-trained MLLM, \model\xspace enables interpretable and flexible TSUR across diverse domains and query types, and extensive experiments demonstrate consistent improvement over both general-purpose LLMs and time-series-specialized baselines.

In summary, our contributions are threefold:
(1) We propose a Patch-level Alignment (PA) module to enforce fine-grained physical consistency between numerical values and visual plots, preventing localized MLLM hallucinations.
(2) We introduce a Discrete Disentangled Interaction (DDI) module to filter shared redundancies and leverage the complementarities of multi-modal signals for robust reasoning.
(3) We conduct extensive experiments on synthetic and real-world data demonstrating the advances of \model\xspace in TSUR.
\section{Related Work}
\label{sec:related}

\subsection{Multi-modal Methods for Classical Time Series Analysis}

Recent studies explore augmenting time series with auxiliary modalities, such as text or images, to enhance classical analysis, particularly forecasting tasks.
One line of research integrates numerical series with textual descriptions~\cite{timecma2025} or exogenous contexts~\cite{camef2025}, typically via dual-branch architectures that encode each modality separately before fusing them for downstream tasks~\cite{timemmd2024,balmtsf2025,timecma2025,dualsg2025,timexl2025}.
With the rise of LLMs, recent efforts enable LLMs to jointly process textual and numerical inputs, by either tokenizing numerical values~\cite{llmtime2023,chattime2025}, or aligning time series representations with LLM-compatible textual embeddings through contrastive learning~\cite{camef2025}, cross-attention mechanisms~\cite{timellm2024,calf2025}, or graph neural networks~\cite{contextalignment2025}.

Another line further incorporates visual modalities derived from time series plots or auxiliary sources.
For instance, Time-VLM~\cite{timevlm2025} visualizes series as heatmaps and generates textual summaries, which are fused with numerical inputs for forecasting.
PIPE~\cite{pipe2025} incorporates satellite imagery as physical cues to enhance typhoon intensity prediction.
these approaches remain mostly task-specific and constrained by fixed output formats, limiting their applicability to flexible reasoning scenarios such as time series understanding and reasoning.

\subsection{Time Series Understanding and Reasoning}

Beyond classical analysis, time series understanding and reasoning (TSUR) emphasizes interpretable analyses of temporal dynamics under flexible natural language queries, leveraging LLMs.
Existing approaches can be broadly classified as numerical-centric, visual-centric, and hybrid paradigms.
Numerical-centric methods represent time series values as sequences of textual or time-series-specific tokens~\cite{merrill2024language,chattime2025,itformer2025,chatts2025}, preserving precision but often struggling with high-level temporal structures due to the text-centric pre-training of LLMs~\cite{tan2024language,vltime2025}.
Visual-centric approaches~\cite{vltime2025,merrill2024language,timemaster2025} reframe time series analysis as a visual question-answering task, transforming the series into plots and leveraging advanced MLLMs~\cite{vltime2025,merrill2024language,timemaster2025}. 
These methods can effectively capture high-level patterns, \eg trend and periodicity, but sacrifice fine-grained numerical details~\cite{visualanomaly2025,zhao2025images}.
Hybrid approaches, \eg GEM~\cite{gem2025}, combine both modalities, but typically rely on simple concatenation and remain domain-specific (\eg ECG), limiting the principled cross-modal alignment and interaction.
Our work addresses these limitations with a generalizable MLLM that integrates numerical precision and visual abstractions via fine-grained alignment and disentangled cross-modal interaction, supporting diverse domains and flexible query types.

\section{Preliminary}

\subsection{Problem Statement}
\label{sec:problem}

Time series understanding and reasoning (TSUR) aim to analyze time series based on a natural language query and generate a textual answer.
Formally, given a set of time series $\mathcal{X}=\{\mathbf{x}_1,\mathbf{x}_2,\dots,\mathbf{x}_{|\mathcal{X}|}\}$, where each $\mathbf{x}_i=(x_{i;1},x_{i;2},\dots,x_{i;T_i})\in\mathbb{R}^{T_i}$ consists of $T_i$ observations, along with textual inputs including a context $C$ and a question $Q$, the objective is to learn a mapping $f:(\mathcal{X},C,Q)\rightarrow A$, producing an answer $A$ that explains or reasons about the relevant information in $\mathcal{X}$ with respect to $Q$.

\subsection{Vector Quantization}
\label{sec:vq}

\emph{Vector Quantization (VQ)}~\cite{vqvae2017} maps continuous features to a finite set of discrete prototypes, enabling compact, structured representations.
Given input features $\mathbf{E}=[\mathbf{e}_1,\dots,\mathbf{e}_N]^\top\in\mathbb{R}^{N\times D}$ and a codebook $\mathcal{C}=\{\mathbf{c}_k\}_{k=1}^K\subseteq \mathbb{R}^D$, each feature is quantized via nearest-neighbor assignment:
$\mathbf{q}_j=\arg\min_{\mathbf{c}_k\in\mathcal{C}}\|\mathbf{e}_j-\mathbf{c}_k\|_2^2$.
To enable gradient-based optimization, VQ employs a straight-through estimator (STE) together with a commitment loss, resulting in the objective $
\mathcal{L}_{\text{vq}} = \|\mathbf{e}_j - \text{sg}[\mathbf{q}_j]\|_2^2$, where $\text{sg}[\cdot]$ denotes the stop-gradient operator.
The codebook is commonly updated using exponential moving average (EMA).

\emph{Residual Vector Quantization (RVQ)}~\cite{rvq2022} extends VQ by sequentially quantizing residuals, increasing representation capacity.
Using $M$ codebooks $\{\mathcal{C}^{(m)}\}_{m=1}^M$, RVQ initializes $\mathbf{r}_j^{(0)}=\mathbf{e}_j$ and iteratively performs $\mathbf{q}_j^{(m)}=\arg\min_{\mathbf{c}\in\mathcal{C}^{(m)}}\|\mathbf{r}_j^{(m-1)}-\mathbf{c}\|_2^2$ and $\mathbf{r}_j^{(m)}=\mathbf{r}_j^{(m-1)}-\mathbf{q}_j^{(m)}$. The final quantized representation is $\mathbf{q}_j=\sum_{m=1}^M\mathbf{q}_j^{(m)}$.



\begin{figure*}[ht]
    \centering
    \includegraphics[width=0.9\linewidth]{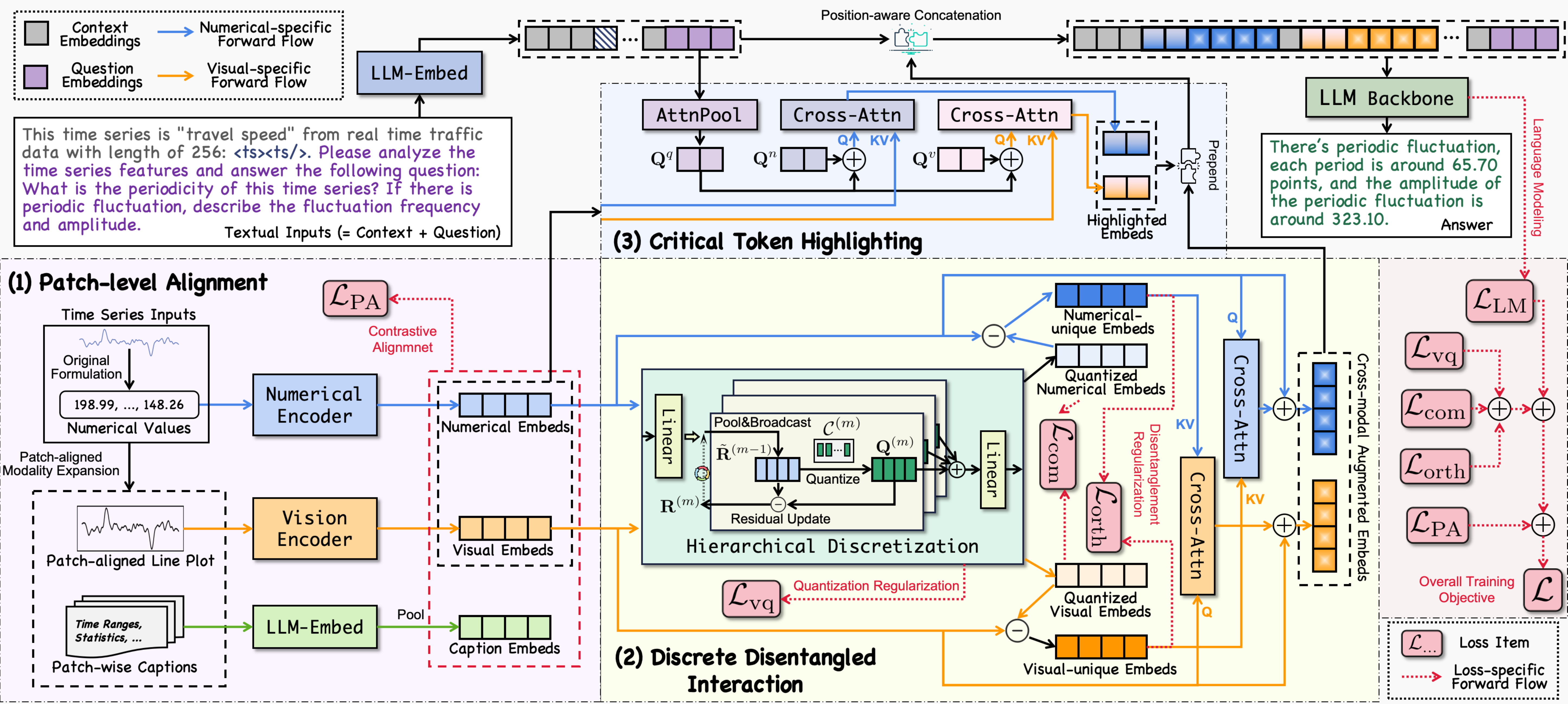}
    \vspace{-5pt}
    \caption{The model architecture of \model.}
    \vspace{-15pt}
    \label{fig:main}
\end{figure*}

\section{Methodology}
\label{sec:method}

\figref{fig:main} illustrates the overall architecture of \model.
It is built upon a pre-trained MLLM, reusing its vision encoder and LLM to perform time series understanding and reasoning (TSUR).
Specifically, \emph{Patch-level Alignment (PA)} first expands each time series as patch-aligned numerical, visual and textual modalities, aligning them in a consistent embedding space.
Based on these aligned token embeddings, \emph{Discrete Disentangled Interaction (DDI)} separates modality-shared and unique semantics for numerical and visual inputs and synergizes their complementarities.
\emph{Critical-token Highlighting (CTH)} then identifies informative, question-relevant tokens from the PA-aligned embeddings and prepends them to the DDI-fused multi-modal sequence.
The resulting multi-modal embeddings are concatenated with the textual input embeddings and decoded by the LLM to generate the output.

\subsection{Patch-level Alignment}
\label{sec:csa}

To establish precise correspondence between numerical and visual modalities and enable accurate MLLM interpretation of numerical inputs, we propose a \emph{Patch-level Alignment (PA)} module that enforces fine-grained, physically-grounded alignment among numerical, visual, and textual modalities.
PA operates at a patch level and consists of three steps: (1) \emph{patch-level modality expansion}, which constructs patch-aligned textual and visual counterparts; (2) \emph{multi-modal encoding}, which embeds each modality independently; and (3) \emph{contrastive alignment}, which employs patch-wise contrastive learning to reduce cross-modal representation gaps.

\subsubsection{Patch-level Modality Expansion}

Following~\cite{patchtst2023}, each time series $\mathbf{x}_i$ can be divided into non-overlapping patches to enable efficient and robust temporal modeling:
$\mathbf{P}^n_i = \text{Patching}(\mathbf{x}_i) \in \mathbb{R}^{\tilde{T}_i\times p^n}$,
where $\tilde{T}_i=[T_i/p^n]$ denotes the number of patches, and $p^n$ is the patch size of time series.
Based on such a patch-level granularity, the numerical modality is then expanded with textual and visual modalities.

\textbf{Patch-aligned Visualization.}
We render $\mathbf{x}_i$ as a line plot with adaptive resolution $\tilde{T}_i \cdot p^v\times p^v$ (width × height) where $p^v$ is the patch size of the vision encoder in the backbone MLLM.
We intentionally omit all decorative plotting elements (\eg titles, axes, ticks, labels, and legends), thereby the resulting image $\mathbf{I}_i$ can be patchified into $\mathbf{P}^v_i \in \mathbb{R}^{\tilde{T}_i\times p^v\times p^v}$, ensuring a one-to-one correspondence with numerical patches $\mathbf{P}^n_i$, and enabling fine-grained numerical-visual alignment.

\textbf{Patch-wise Captioning.}
For numerical-text alignment, prior studies rely on time-series-agnostic texts, \eg task prompts~\cite{contextalignment2025}, external contexts~\cite{camef2025}, general word embeddings~\cite{timellm2024}, which fail to capture nuanced physical variations of time series (\eg amplitude, frequency), restricting semantic fidelity. 
We instead construct patch-wise captions grounded in numerical values.
For each numerical patch $\mathbf{p}^n_{i;j}$, a structured caption $s_{i;j}$ consists of timestamp ranges and value statistics (\eg max, min, mean, standard deviation), yielding a set of patch-wise captions $S_i = \{s_{i;1}, \dots, s_{i;\tilde{T}_i}\}$.

\subsubsection{Multi-modal Encoding}
Given expanded multi-modal inputs $\{\mathbf{x}_i,\mathbf{I_i},S_i\}_{i=1}^{|\mathcal{X}|}$, numerical, visual, and textual modalities are encoded independently into continuous patch-level embeddings.
Each instance is processed independently to accommodate variable sequence lengths, while cross-instance dependencies are captured implicitly by the backbone LLM.

\textbf{Numerical Encoding.} We adopt a lightweight time series encoder consisting of linear projection, positional embedding, and transformer blocks:
\begin{equation}
\begin{aligned}
\mathbf{P}^n_i &= [\mathbf{p}^n_{i;1}, \dots, \mathbf{p}^n_{i;\tilde{T}_i}]^\top = \text{Patching}(\mathbf{x}_i) \in \mathbb{R}^{\tilde{T}_i\times p^n}, \\
\mathbf{B}^n_i &= \text{LinearBlocks}(\text{Concat}(\mathbf{P}^n_i,\text{PosEmbed}(\mathbf{P}^n_i))), \\
\mathbf{E}^n_i &= \text{TransformerBlocks}(\mathbf{B}^n_i)\in \mathbb{R}^{\tilde{T}_i\times D}.
\end{aligned}
\end{equation}

\textbf{Visual Encoding.} The time series line plot is processed by the pre-trained vision encoder of the backbone MLLM:
$\mathbf{E}^v_i = [\mathbf{e}^v_{i;1},\dots,\mathbf{e}^v_{i;\tilde{T}_i}]^\top=\text{VisionEncoder}(\mathbf{I}_i)\in \mathbb{R}^{\tilde{T}_i\times D}$.

\textbf{Caption Encoding.}
To encode the patch-wise caption $s_{i;j}$, we reuse the tokenizer and the input embedding layer of the backbone LLM. 
Then a mean pooling layer is applied to obtain patch-specific caption embeddings. 
This process is formulated as $\mathbf{e}^s_{i;j} = \text{MeanPool}(\text{LLM-Embed}(s_{i;j}))$, and $\mathbf{E}^s_i = [\mathbf{e}^s_{i;1},\dots,\mathbf{e}^s_{i;\tilde{T}_i}]^\top \in \mathbb{R}^{\tilde{T}_i\times D}$ is the stacked tensor.

\subsubsection{Contrastive Alignment}

To strengthen semantic correspondence across modalities, we adopt a patch-wise contrastive learning objective.
Numerical patch embeddings serve as anchors, while patch-aligned visual and textual caption embeddings are positives; all other patches within the same instance act as negatives.
The \emph{numerical–visual} alignment loss is defined with an InfoNCE objective~\cite{oord2018representation}:
\begin{equation}
\mathcal{L}^{n\text{-}v}_{\text{align}} = - \sum_{i,j}\log\frac{\exp(\text{sim}(\mathbf{e}^n_{i;j}, \text{sg}(\mathbf{e}^v_{i;j}))/\tau)}{\sum_{j^\prime}\exp(\text{sim}(\mathbf{e}^n_{i;j}, \text{sg}(\mathbf{e}^v_{i;j^\prime}))/\tau)},
\end{equation}
where $\text{sim}$ denotes cosine similarity, $\text{sg}$ denotes the stop-gradient operator, $\tau$ is a temperature parameter, and $(i,j^\prime)$ indexes all visual tokens within the $i$-th time series instance.
The \emph{numerical–caption} alignment loss $\mathcal{L}^{n\text{-}s}_{\text{align}}$ is computed analogously.
We stop the gradients of encoded visual and textual embeddings to optimize the numerical encoder only.
The overall alignment objective is:
$\mathcal{L}_{\text{PA}} = \mathcal{L}^{n\text{-}v}_{\text{align}} + \mathcal{L}^{n\text{-}s}_{\text{align}}$.

\subsection{Discrete Disentangled Interaction}
\label{sec:dci}

Building upon the aligned features, effective TSUR requires consolidating complementary semantics from numerical and visual modalities.
We propose a \emph{Discrete Disentangled Interaction (DDI)} module that performs modality disentanglement in a discrete latent space to enforce compact, consistent representations, and then integrates the isolated cross-modal unique signals.
DDI consists of two stages:
(1) \emph{discrete modality disentanglement}, which extracts compact modality-common semantics via hierarchical vector quantization and derives modality-unique signals by residual decomposition; and
(2) \emph{unique-centric interaction}, which synergizes the unique signals via cross-attention.

\subsubsection{Discrete Modality Disentanglement}

A straightforward approach to identify modality-unique semantics is to apply conventional disentanglement techniques~\cite{decalign2025}, which use separate projection heads to extract modality-common and modality-specific representations, alongside regularization encouraging high correlation among common components and low correlation with unique components.
However, existing methods~\cite{xia2023achieving,huang2025enhancing} operate in continuous embedding spaces, which are unbounded and prone to capturing redundant or noisy correlations.
In contrast, we perform disentanglement in a \emph{discrete latent space} using vector quantization (VQ), which enforces compactness and consistency for modality-common semantics.
Given numerical and visual token embeddings $\mathbf{E}_i^n,\mathbf{E}_i^v\in\mathbb{R}^{\tilde{T}_i\times D}$, we employ a shared hierarchical discretization block to extract modality-common representations $\mathbf{Z}_i^n$ and $\mathbf{Z}_i^v$.
The modality-unique components are then obtained by residual decomposition:
\begin{equation}
\mathbf{U}_i^n=\mathbf{E}_i^n-\mathbf{Z}_i^n,\quad
\mathbf{U}_i^v=\mathbf{E}_i^v-\mathbf{Z}_i^v.
\label{equ:unique}
\end{equation}


\textbf{Hierarchical Discretization.}
To capture the multi-scale time series semantics, we propose a hierarchical RVQ strategy featured by $M$ hierarchical codebooks $\{\mathcal{C}^{(m)}\}_{m=1}^M$, where the $m$-th codebook $\mathcal{C}^{(m)}$ contains $K \cdot 2^{m-1}$ codes and operates at a temporal resolution downsampled by $2^{M-m}$ tokens.
Specifically, we first follow~\cite{yu2022vectorquantized} and project continuous embeddings $\mathbf{E}_i^n\in\mathbb{R}^{\tilde{T}_i\times D}$ to a lower-dimensional space via $\tilde{\mathbf{E}}_i^n=\text{Linear}(\mathbf{E}_i^n)\in\mathbb{R}^{\tilde{T}_i\times d}$,
which increases the density of codes and reduces the quantization error.
At each quantization stage $m$, given the residual token sequence $\mathbf{R}_i^{n;(m-1)}=[\mathbf{r}_{i;1}^{n;(m-1)},\dots,\mathbf{r}_{i;\tilde{T}_i}^{n;(m-1)}]^\top\in\mathbb{R}^{\tilde{T}_i\times d}$ from the last round, we partition it into non-overlapping segments of length $W=2^{M-m}$, and compute segment-level representations by average pooling over the tokens within each segment.
These pooled representations are then broadcasted back to all token positions in the corresponding segment.
This process can be formulated as $\tilde{\mathbf{r}}_{i;j}^{n;(m-1)}= \frac{1}{W}\sum_{b=j^*}^{j^*+W}\mathbf{r}_{i;b}^{n;(m-1)}$, where $j^*=[(j-1)/W]$ denotes the segment index of the $j$-th token.
Afterwards, we follow the steps illustrated in \secref{sec:vq}, iteratively quantizing the residual features, from coarse to fine scales:
\begin{equation}
\begin{aligned}
\mathbf{q}_{i;j}^{n;(m)} &= \arg\min_{\mathbf{c} \in \mathcal{C}^{(m)}} \|\tilde{\mathbf{r}}_{i;j}^{n;(m-1)}- \mathbf{c}\|_2^2,\\
\mathbf{r}_{i;j}^{n;(m)} &= \mathbf{r}_{i;j}^{n;(m-1)} - \mathbf{q}_{i;j}^{n;(m)},
\end{aligned}
\end{equation}
where the initial residual token embeddings are $\mathbf{R}_i^{n;(0)}=\tilde{\mathbf{E}}_i^n$.
The final discretized features are accumulated by $\mathbf{q}_{i;j}^{n} = \sum_{m=1}^M \mathbf{q}_{i;j}^{n;(m)}$.
After quantization, the discrete representations should be projected back to $D$-dimensional space via $\mathbf{z}_{i;j}^{n}=\text{Linear}(\mathbf{q}_{i;j}^{n})\in \mathbb{R}^D$,
which serve as the modality-common tokens, and constitute $\mathbf{Z}_i^{n}\in \mathbb{R}^{\tilde{T}_i\times d}$.
The modality-common token embeddings of visual time series $\mathbf{Z}_i^{v}$ are derived in a similar way. 
The qunatization process is supervised the $\mathcal{L}_{\text{vq}}$ illustrated in \secref{sec:vq}.

\textbf{Disentanglement Regularization.}
To effectively decouple modality-common and modality-unique information, we introduce two regularization terms.
First, to align modality-common semantics, we apply a bidirectional contrastive objective between quantized numerical and visual token embeddings: $\mathcal{L}_{\text{com}}=\frac{1}{2}(\mathcal{L}^{n\text{-}v}_{\text{com}}+\mathcal{L}^{v\text{-}n}_{\text{com}})$, where 
\begin{equation}
\mathcal{L}^{n\text{-}v}_{\text{com}}= - \sum_{i,j}\log\frac{\exp(\text{sim}(\mathbf{z}^n_{i;j}, \mathbf{z}^v_{i;j})/\tau)}{\sum_{j^\prime}\exp(\text{sim}(\mathbf{z}^n_{i;j}, \mathbf{z}^v_{i;j^\prime})/\tau)}.
\end{equation}
To suppress leakage of shared semantics into modality-unique components, we enforce orthogonality between the common and unique token embeddings:
\begin{equation}
\mathcal{L}_{\text{orth}}= \frac{1}{2}\sum_{i,j}(|\text{sim}(\mathbf{z}^n_{i;j}, \mathbf{u}^n_{i;j})|+|\text{sim}(\mathbf{z}^v_{i;j}, \mathbf{u}^v_{i;j})|).
\end{equation}
The overall objective for DDI is:
$\mathcal{L}_{\text{DDI}}
= \mathcal{L}_{\text{vq}} + \alpha\cdot\mathcal{L}_{\text{com}} + \beta\cdot\mathcal{L}_{\text{orth}}$, where $\alpha,\beta$ control the regularization strengths.

\subsubsection{Unique-centric Interaction}

After disentanglement, modality-unique semantics are integrated via cross-modal attention.
Each modality’s original tokens are augmented by attending to the unique signals (derived by \equref{equ:unique}) from the opposite modality:
\begin{align}
\overline{\mathbf{E}}^n_i &= \mathbf{E}^n_i +\text{CrossAttn}(\mathbf{E}^n_i,\mathbf{U}^v_i,\mathbf{U}^v_i), \\
\overline{\mathbf{E}}^v_i &= \mathbf{E}^v_i +\text{CrossAttn}(\mathbf{E}^v_i,\mathbf{U}^n_i,\mathbf{U}^n_i).
\end{align}
This design enables fine-grained numerical dynamics and high-level visual patterns to complement each other.

\subsection{Critical-token Highlighting}
\label{sec:cth}

Although PA and DDI yield aligned and complementary representations, directly feeding them into the LLM may dilute question-relevant or inherently important information, limiting generalization.
We propose a \emph{Critical-token Highlighting (CTH)} module, which retains all original tokens and prepends several informative tokens to explicitly emphasize key signals, conditioned on both the question-aware and modality-intrinsic saliency.
Specifically, the question $Q$ is first tokenized and embedded as $\mathbf{E}^q = \text{LLM-Embed}(Q)\in \mathbb{R}^{N^q\times D}$.
Given question embeddings $\mathbf{E}^q$, and continuous numerical/visual embeddings $\mathbf{E}_i^o\in \mathbb{R}^{\tilde{T}_i \times D}$ ($o\in\{n,v\}$), CTH employs two parallel cross-attention branches.

\textbf{Question-conditioned Branch.}
We first compress the variable-length question embeddings into $H$ learnable query embeddings $\mathbf{Q}^{q}$ via an attention-based pooling layer, and then use the compressed question-aware queries to attend over the modality tokens, which can be formulated as:
\begin{equation}
\begin{aligned}
\tilde{\mathbf{Q}}^{q} &= \text{CrossAttn}(\mathbf{Q}^{q},\mathbf{E}^q,\mathbf{E}^q),\\
\mathbf{H}^{o;q}_i = &\text{CrossAttn}(\tilde{\mathbf{Q}}^{q},\mathbf{E}_i^o,\mathbf{E}_i^o),\ o\in\{n,v\}.
\end{aligned}
\end{equation}

\textbf{Modality-intrinsic Branch.}
In parallel, we introduce another set of $H$ learnable queries $\mathbf{Q}^{o} \in \mathbb{R}^{H \times D}$ to identify modality-intrinsic patterns:
\begin{equation}
\mathbf{H}^{o;s}_i = \text{CrossAttn}(\mathbf{Q}^{o},\mathbf{E}_i^o,\mathbf{E}_i^o),\ o\in\{n,v\}.
\end{equation}

The highlighted token embeddings are fused as $\mathbf{H}_i^o = \mathbf{H}^{o;q}_i + \mathbf{H}^{o;s}_i$ and prepended to the modality token sequence via $\hat{\mathbf{E}}^o_i = \text{Concat}(\mathbf{H}_i^o,\overline{\mathbf{E}}^o_i)\in\mathbb{R}^{(H+\tilde{T}_i)\times D},\ o\in\{n,v\}$. 
This results in holistic and question-aware modality representations for LLM understanding and reasoning.

\begin{table*}[!t]
\centering
\caption{Performance (\%) on understanding tasks. The light-blue background denotes results from general-purpose LLMs/MLLMs accessed via APIs, while the light-red background denotes results from time-series-specific MLLMs.}
\vspace{-5pt}
\scriptsize
\begin{tabular}{c | c |c |c c c c c c c c  |  c c | c c}
\toprule
\multirow{2}{*}{\textbf{Dataset}} & \multirow{2}{*}{\textbf{Type}} & \multirow{2}{*}{\textbf{Model}} & \multicolumn{2}{c}{\textbf{Noise}}& \multicolumn{2}{c}{\textbf{Local}}& \multicolumn{2}{c}{\textbf{Season}}& \multicolumn{2}{c}{\textbf{Trend}}& \multicolumn{1}{c}{\textbf{Corr.}}& \multicolumn{1}{c}{\textbf{Clus.}}& \multicolumn{2}{c}{\textbf{Overall}} \\
\cmidrule(lr){4-13}\cmidrule(lr){14-15}
& & & \textbf{Cate.} & \textbf{Num.} & \textbf{Cate.} & \textbf{Num.} & \textbf{Cate.} & \textbf{Num.} & \textbf{Cate.} & \textbf{Num.} & \textbf{Cate.} & \textbf{Cate.} & \textbf{Cate.} & \textbf{Num.} \\

\midrule
\multirow{19}{*}{\rotatebox{90}{\textbf{Synthetic}}} & \multirow{9}{*}{\rotatebox{90}{\textbf{Numerical}}} & \cellcolor{blue!10}\textbf{DeepSeek-V3.2} & \cellcolor{blue!10}74.39 & \cellcolor{blue!10}47.15 & \cellcolor{blue!10}31.70 & \cellcolor{blue!10}23.13 & \cellcolor{blue!10}78.35 & \cellcolor{blue!10}67.16 & \cellcolor{blue!10}78.57 & \cellcolor{blue!10}85.04 & \cellcolor{blue!10}36.64 & \cellcolor{blue!10}30.20 & \cellcolor{blue!10}43.81 & \cellcolor{blue!10}37.74 \\
 &  & \cellcolor{blue!10}\textbf{GPT-4o} & \cellcolor{blue!10}56.10 & \cellcolor{blue!10}43.21 & \cellcolor{blue!10}14.77 & \cellcolor{blue!10}9.72 & \cellcolor{blue!10}88.66 & \cellcolor{blue!10}55.16 & \cellcolor{blue!10}72.62 & \cellcolor{blue!10}83.80 & \cellcolor{blue!10}34.81 & \cellcolor{blue!10}30.09 & \cellcolor{blue!10}38.36 & \cellcolor{blue!10}26.81 \\
 &  & \cellcolor{blue!10}\textbf{Qwen3} & \cellcolor{blue!10}63.41 & \cellcolor{blue!10}40.64 & \cellcolor{blue!10}24.23 & \cellcolor{blue!10}15.69 & \cellcolor{blue!10}85.57 & \cellcolor{blue!10}51.25 & \cellcolor{blue!10}66.67 & \cellcolor{blue!10}83.48 & \cellcolor{blue!10}30.11 & \cellcolor{blue!10}29.36 & \cellcolor{blue!10}39.01 & \cellcolor{blue!10}30.67 \\
 &  & \cellcolor{blue!10}\textbf{GPT-5.2} & \cellcolor{blue!10}73.17 & \cellcolor{blue!10}43.03 & \cellcolor{blue!10}41.41 & \cellcolor{blue!10}34.88 & \cellcolor{blue!10}96.91 & \cellcolor{blue!10}72.09 & \cellcolor{blue!10}78.57 & \cellcolor{blue!10}93.01 & \cellcolor{blue!10}47.20 & \cellcolor{blue!10}37.27 & \cellcolor{blue!10}52.27 & \cellcolor{blue!10}47.88 \\
 &  & \cellcolor{blue!10}\textbf{Gemini 3 Pro} & \cellcolor{blue!10}84.15 & \cellcolor{blue!10}59.67 & \cellcolor{blue!10}53.12 & \cellcolor{blue!10}45.82 & \cellcolor{blue!10}95.88 & \cellcolor{blue!10}84.10 & \cellcolor{blue!10}76.19 & \cellcolor{blue!10}83.26 & \cellcolor{blue!10}54.12 & \cellcolor{blue!10}49.16 & \cellcolor{blue!10}60.48 & \cellcolor{blue!10}55.63 \\
 &  & \cellcolor{red!10}\textbf{ChatTime} & \cellcolor{red!10}84.15 & \cellcolor{red!10}13.74 & \cellcolor{red!10}67.14 & \cellcolor{red!10}35.49 & \cellcolor{red!10}95.88 & \cellcolor{red!10}49.08 & \cellcolor{red!10}91.67 & \cellcolor{red!10}12.65 & \cellcolor{red!10}71.95 & \cellcolor{red!10}69.84 & \cellcolor{red!10}74.94 & \cellcolor{red!10}32.21 \\
 &  & \cellcolor{red!10}\textbf{ChatTS} & \cellcolor{red!10}93.90 & \cellcolor{red!10}47.20 & \cellcolor{red!10}\underline{\emph{84.38}} & \cellcolor{red!10}\underline{\emph{76.02}} & \cellcolor{red!10}\textbf{100.00} & \cellcolor{red!10}86.64 & \cellcolor{red!10}\textbf{96.43} & \cellcolor{red!10}\underline{\emph{96.34}} & \cellcolor{red!10}81.07 & \cellcolor{red!10}\underline{\emph{79.27}} & \cellcolor{red!10}\underline{\emph{85.32}} & \cellcolor{red!10}\underline{\emph{79.54}} \\
 &  & \cellcolor{red!10}\textbf{ITFormer} & \cellcolor{red!10}85.37 & \cellcolor{red!10}23.32 & \cellcolor{red!10}20.69 & \cellcolor{red!10}13.71 & \cellcolor{red!10}71.13 & \cellcolor{red!10}26.92 & \cellcolor{red!10}70.24 & \cellcolor{red!10}94.19 & \cellcolor{red!10}56.66 & \cellcolor{red!10}47.50 & \cellcolor{red!10}50.32 & \cellcolor{red!10}28.59 \\
 &  & \cellcolor{red!10}\textbf{InstructTime} & \cellcolor{red!10}79.27 & \cellcolor{red!10}19.34 & \cellcolor{red!10}22.32 & \cellcolor{red!10}15.24 & \cellcolor{red!10}70.10 & \cellcolor{red!10}11.23 & \cellcolor{red!10}72.62 & \cellcolor{red!10}96.27 & \cellcolor{red!10}52.29 & \cellcolor{red!10}48.06 & \cellcolor{red!10}49.34 & \cellcolor{red!10}28.60 \\
\cmidrule(lr){2-13}
\cmidrule(lr){14-15}
 & \multirow{4}{*}{\rotatebox{90}{\textbf{Visual}}} & \cellcolor{blue!10}\textbf{GPT-4o} & \cellcolor{blue!10}64.63 & \cellcolor{blue!10}31.72 & \cellcolor{blue!10}34.39 & \cellcolor{blue!10}26.97 & \cellcolor{blue!10}87.63 & \cellcolor{blue!10}65.01 & \cellcolor{blue!10}50.00 & \cellcolor{blue!10}48.23 & \cellcolor{blue!10}39.54 & \cellcolor{blue!10}43.90 & \cellcolor{blue!10}46.59 & \cellcolor{blue!10}33.79 \\
 &  & \cellcolor{blue!10}\textbf{Qwen3-VL} & \cellcolor{blue!10}53.66 & \cellcolor{blue!10}18.92 & \cellcolor{blue!10}34.37 & \cellcolor{blue!10}26.31 & \cellcolor{blue!10}90.72 & \cellcolor{blue!10}49.71 & \cellcolor{blue!10}80.95 & \cellcolor{blue!10}49.08 & \cellcolor{blue!10}38.81 & \cellcolor{blue!10}41.70 & \cellcolor{blue!10}47.71 & \cellcolor{blue!10}31.92 \\
 &  & \cellcolor{blue!10}\textbf{GPT-5.2} & \cellcolor{blue!10}75.61 & \cellcolor{blue!10}66.64 & \cellcolor{blue!10}52.28 & \cellcolor{blue!10}44.12 & \cellcolor{blue!10}96.91 & \cellcolor{blue!10}74.54 & \cellcolor{blue!10}78.57 & \cellcolor{blue!10}84.61 & \cellcolor{blue!10}61.03 & \cellcolor{blue!10}52.49 & \cellcolor{blue!10}62.55 & \cellcolor{blue!10}54.07 \\
 &  & \cellcolor{blue!10}\textbf{Gemini 3 Pro} & \cellcolor{blue!10}91.46 & \cellcolor{blue!10}60.52 & \cellcolor{blue!10}53.84 & \cellcolor{blue!10}46.49 & \cellcolor{blue!10}94.85 & \cellcolor{blue!10}\underline{\emph{89.58}} & \cellcolor{blue!10}80.95 & \cellcolor{blue!10}79.89 & \cellcolor{blue!10}65.39 & \cellcolor{blue!10}55.38 & \cellcolor{blue!10}66.01 & \cellcolor{blue!10}56.06 \\
\cmidrule(lr){2-13}
\cmidrule(lr){14-15}
 & \multirow{6}{*}{\rotatebox{90}{\textbf{Num.+Visual}}} & \cellcolor{blue!10}\textbf{GPT-4o} & \cellcolor{blue!10}58.54 & \cellcolor{blue!10}40.02 & \cellcolor{blue!10}24.60 & \cellcolor{blue!10}19.20 & \cellcolor{blue!10}93.81 & \cellcolor{blue!10}69.00 & \cellcolor{blue!10}46.43 & \cellcolor{blue!10}82.19 & \cellcolor{blue!10}33.72 & \cellcolor{blue!10}37.38 & \cellcolor{blue!10}40.89 & \cellcolor{blue!10}34.41 \\
 &  & \cellcolor{blue!10}\textbf{Qwen3-VL} & \cellcolor{blue!10}48.78 & \cellcolor{blue!10}27.21 & \cellcolor{blue!10}23.41 & \cellcolor{blue!10}18.34 & \cellcolor{blue!10}86.60 & \cellcolor{blue!10}54.67 & \cellcolor{blue!10}83.33 & \cellcolor{blue!10}70.71 & \cellcolor{blue!10}36.64 & \cellcolor{blue!10}36.91 & \cellcolor{blue!10}42.73 & \cellcolor{blue!10}30.40 \\
 &  & \cellcolor{blue!10}\textbf{GPT-5.2} & \cellcolor{blue!10}73.17 & \cellcolor{blue!10}61.05 & \cellcolor{blue!10}43.25 & \cellcolor{blue!10}36.42 & \cellcolor{blue!10}98.97 & \cellcolor{blue!10}79.38 & \cellcolor{blue!10}75.00 & \cellcolor{blue!10}93.61 & \cellcolor{blue!10}59.21 & \cellcolor{blue!10}48.32 & \cellcolor{blue!10}58.58 & \cellcolor{blue!10}50.27 \\
 &  & \cellcolor{blue!10}\textbf{Gemini 3 Pro} & \cellcolor{blue!10}86.59 & \cellcolor{blue!10}\underline{\emph{70.51}} & \cellcolor{blue!10}53.92 & \cellcolor{blue!10}46.88 & \cellcolor{blue!10}97.94 & \cellcolor{blue!10}86.52 & \cellcolor{blue!10}76.19 & \cellcolor{blue!10}83.88 & \cellcolor{blue!10}63.57 & \cellcolor{blue!10}54.72 & \cellcolor{blue!10}64.90 & \cellcolor{blue!10}56.96 \\
 &  & \cellcolor{red!10}\textbf{GEM} & \cellcolor{red!10}\underline{\emph{96.34}} & \cellcolor{red!10}56.25 & \cellcolor{red!10}80.81 & \cellcolor{red!10}70.08 & \cellcolor{red!10}\textbf{100.00} & \cellcolor{red!10}85.01 & \cellcolor{red!10}91.67 & \cellcolor{red!10}95.77 & \cellcolor{red!10}\underline{\emph{82.15}} & \cellcolor{red!10}78.45 & \cellcolor{red!10}84.36 & \cellcolor{red!10}75.27 \\
 &  & \cellcolor{red!10}\textbf{\model} & \cellcolor{red!10}\textbf{98.78} & \cellcolor{red!10}\textbf{78.39} & \cellcolor{red!10}\textbf{91.45} & \cellcolor{red!10}\textbf{84.73} & \cellcolor{red!10}\textbf{100.00} & \cellcolor{red!10}\textbf{93.27} & \cellcolor{red!10}\underline{\emph{94.05}} & \cellcolor{red!10}\textbf{96.60} & \cellcolor{red!10}\textbf{86.89} & \cellcolor{red!10}\textbf{84.35} & \cellcolor{red!10}\textbf{89.99} & \cellcolor{red!10}\textbf{87.22} \\

\midrule
\multirow{19}{*}{\rotatebox{90}{\textbf{Real-world}}} & \multirow{9}{*}{\rotatebox{90}{\textbf{Numerical}}} & \cellcolor{blue!10}\textbf{DeepSeek-V3.2} & \cellcolor{blue!10}85.71 & \cellcolor{blue!10}29.30 & \cellcolor{blue!10}60.72 & \cellcolor{blue!10}46.14 & \cellcolor{blue!10}81.08 & \cellcolor{blue!10}0.00 & \cellcolor{blue!10}65.85 & \cellcolor{blue!10}88.09 & \cellcolor{blue!10}61.11 & \cellcolor{blue!10}51.73 & \cellcolor{blue!10}63.86 & \cellcolor{blue!10}50.35 \\
 &  & \cellcolor{blue!10}\textbf{GPT-4o} & \cellcolor{blue!10}95.24 & \cellcolor{blue!10}19.97 & \cellcolor{blue!10}32.15 & \cellcolor{blue!10}24.16 & \cellcolor{blue!10}78.38 & \cellcolor{blue!10}0.00 & \cellcolor{blue!10}70.73 & \cellcolor{blue!10}84.49 & \cellcolor{blue!10}57.03 & \cellcolor{blue!10}46.12 & \cellcolor{blue!10}54.80 & \cellcolor{blue!10}34.86 \\
 &  & \cellcolor{blue!10}\textbf{Qwen3} & \cellcolor{blue!10}78.57 & \cellcolor{blue!10}37.44 & \cellcolor{blue!10}46.78 & \cellcolor{blue!10}29.73 & \cellcolor{blue!10}56.76 & \cellcolor{blue!10}49.22 & \cellcolor{blue!10}56.10 & \cellcolor{blue!10}81.31 & \cellcolor{blue!10}76.67 & \cellcolor{blue!10}44.31 & \cellcolor{blue!10}55.54 & \cellcolor{blue!10}40.49 \\
 &  & \cellcolor{blue!10}\textbf{GPT-5.2} & \cellcolor{blue!10}92.86 & \cellcolor{blue!10}31.88 & \cellcolor{blue!10}57.86 & \cellcolor{blue!10}45.06 & \cellcolor{blue!10}\underline{\emph{97.30}} & \cellcolor{blue!10}0.00 & \cellcolor{blue!10}60.98 & \cellcolor{blue!10}96.02 & \cellcolor{blue!10}81.11 & \cellcolor{blue!10}56.90 & \cellcolor{blue!10}69.47 & \cellcolor{blue!10}52.33 \\
 &  & \cellcolor{blue!10}\textbf{Gemini 3 Pro} & \cellcolor{blue!10}92.86 & \cellcolor{blue!10}24.76 & \cellcolor{blue!10}86.07 & \cellcolor{blue!10}72.08 & \cellcolor{blue!10}91.89 & \cellcolor{blue!10}70.27 & \cellcolor{blue!10}56.10 & \cellcolor{blue!10}81.16 & \cellcolor{blue!10}84.08 & \cellcolor{blue!10}76.88 & \cellcolor{blue!10}80.25 & \cellcolor{blue!10}67.36 \\
 &  & \cellcolor{red!10}\textbf{ChatTime} & \cellcolor{red!10}50.00 & \cellcolor{red!10}2.92 & \cellcolor{red!10}83.21 & \cellcolor{red!10}48.55 & \cellcolor{red!10}86.49 & \cellcolor{red!10}43.71 & \cellcolor{red!10}\textbf{97.56} & \cellcolor{red!10}5.97 & \cellcolor{red!10}85.56 & \cellcolor{red!10}54.51 & \cellcolor{red!10}75.36 & \cellcolor{red!10}34.75 \\
 &  & \cellcolor{red!10}\textbf{ChatTS} & \cellcolor{red!10}88.10 & \cellcolor{red!10}\underline{\emph{47.09}} & \cellcolor{red!10}86.78 & \cellcolor{red!10}78.40 & \cellcolor{red!10}\underline{\emph{97.30}} & \cellcolor{red!10}\underline{\emph{84.60}} & \cellcolor{red!10}\underline{\emph{92.68}} & \cellcolor{red!10}90.01 & \cellcolor{red!10}68.52 & \cellcolor{red!10}62.94 & \cellcolor{red!10}82.39 & \cellcolor{red!10}77.01 \\
 &  & \cellcolor{red!10}\textbf{ITFormer} & \cellcolor{red!10}69.05 & \cellcolor{red!10}19.85 & \cellcolor{red!10}40.12 & \cellcolor{red!10}28.63 & \cellcolor{red!10}78.38 & \cellcolor{red!10}44.59 & \cellcolor{red!10}70.73 & \cellcolor{red!10}90.84 & \cellcolor{red!10}43.70 & \cellcolor{red!10}40.58 & \cellcolor{red!10}52.04 & \cellcolor{red!10}38.74 \\
 &  & \cellcolor{red!10}\textbf{InstructTime} & \cellcolor{red!10}54.76 & \cellcolor{red!10}11.22 & \cellcolor{red!10}32.73 & \cellcolor{red!10}24.67 & \cellcolor{red!10}\underline{\emph{97.30}} & \cellcolor{red!10}0.00 & \cellcolor{red!10}75.61 & \cellcolor{red!10}\underline{\emph{96.04}} & \cellcolor{red!10}52.59 & \cellcolor{red!10}30.89 & \cellcolor{red!10}51.53 & \cellcolor{red!10}36.16 \\
\cmidrule(lr){2-13}
\cmidrule(lr){14-15}
 & \multirow{4}{*}{\rotatebox{90}{\textbf{Visual}}} & \cellcolor{blue!10}\textbf{GPT-4o} & \cellcolor{blue!10}\textbf{97.62} & \cellcolor{blue!10}20.10 & \cellcolor{blue!10}64.41 & \cellcolor{blue!10}53.05 & \cellcolor{blue!10}94.59 & \cellcolor{blue!10}0.00 & \cellcolor{blue!10}34.15 & \cellcolor{blue!10}68.48 & \cellcolor{blue!10}55.18 & \cellcolor{blue!10}62.75 & \cellcolor{blue!10}64.22 & \cellcolor{blue!10}50.17 \\
 &  & \cellcolor{blue!10}\textbf{Qwen3-VL} & \cellcolor{blue!10}\textbf{97.62} & \cellcolor{blue!10}30.93 & \cellcolor{blue!10}63.81 & \cellcolor{blue!10}50.41 & \cellcolor{blue!10}59.46 & \cellcolor{blue!10}20.54 & \cellcolor{blue!10}80.49 & \cellcolor{blue!10}53.75 & \cellcolor{blue!10}54.45 & \cellcolor{blue!10}61.80 & \cellcolor{blue!10}66.97 & \cellcolor{blue!10}47.60 \\
 &  & \cellcolor{blue!10}\textbf{GPT-5.2} & \cellcolor{blue!10}95.24 & \cellcolor{blue!10}43.38 & \cellcolor{blue!10}83.21 & \cellcolor{blue!10}71.77 & \cellcolor{blue!10}\underline{\emph{97.30}} & \cellcolor{blue!10}0.00 & \cellcolor{blue!10}65.85 & \cellcolor{blue!10}90.66 & \cellcolor{blue!10}87.78 & \cellcolor{blue!10}75.43 & \cellcolor{blue!10}82.33 & \cellcolor{blue!10}70.39 \\
 &  & \cellcolor{blue!10}\textbf{Gemini 3 Pro} & \cellcolor{blue!10}\textbf{97.62} & \cellcolor{blue!10}43.27 & \cellcolor{blue!10}86.31 & \cellcolor{blue!10}73.28 & \cellcolor{blue!10}\underline{\emph{97.30}} & \cellcolor{blue!10}30.28 & \cellcolor{blue!10}75.61 & \cellcolor{blue!10}90.42 & \cellcolor{blue!10}\textbf{92.96} & \cellcolor{blue!10}\underline{\emph{81.23}} & \cellcolor{blue!10}\underline{\emph{86.58}} & \cellcolor{blue!10}71.95 \\
\cmidrule(lr){2-13}
\cmidrule(lr){14-15}
 & \multirow{6}{*}{\rotatebox{90}{\textbf{Num.+Visual}}} & \cellcolor{blue!10}\textbf{GPT-4o} & \cellcolor{blue!10}95.24 & \cellcolor{blue!10}28.12 & \cellcolor{blue!10}44.65 & \cellcolor{blue!10}36.79 & \cellcolor{blue!10}91.89 & \cellcolor{blue!10}0.00 & \cellcolor{blue!10}34.15 & \cellcolor{blue!10}86.28 & \cellcolor{blue!10}60.37 & \cellcolor{blue!10}51.24 & \cellcolor{blue!10}56.00 & \cellcolor{blue!10}43.63 \\
 &  & \cellcolor{blue!10}\textbf{Qwen3-VL} & \cellcolor{blue!10}95.24 & \cellcolor{blue!10}29.21 & \cellcolor{blue!10}40.35 & \cellcolor{blue!10}30.80 & \cellcolor{blue!10}72.97 & \cellcolor{blue!10}42.70 & \cellcolor{blue!10}80.49 & \cellcolor{blue!10}67.15 & \cellcolor{blue!10}66.29 & \cellcolor{blue!10}48.39 & \cellcolor{blue!10}59.97 & \cellcolor{blue!10}37.30 \\
 &  & \cellcolor{blue!10}\textbf{GPT-5.2} & \cellcolor{blue!10}88.10 & \cellcolor{blue!10}43.86 & \cellcolor{blue!10}82.98 & \cellcolor{blue!10}71.40 & \cellcolor{blue!10}\underline{\emph{97.30}} & \cellcolor{blue!10}0.00 & \cellcolor{blue!10}65.85 & \cellcolor{blue!10}\textbf{96.14} & \cellcolor{blue!10}87.41 & \cellcolor{blue!10}68.44 & \cellcolor{blue!10}79.44 & \cellcolor{blue!10}71.03 \\
 &  & \cellcolor{blue!10}\textbf{Gemini 3 Pro} & \cellcolor{blue!10}\textbf{97.62} & \cellcolor{blue!10}46.32 & \cellcolor{blue!10}86.31 & \cellcolor{blue!10}72.76 & \cellcolor{blue!10}\underline{\emph{97.30}} & \cellcolor{blue!10}68.75 & \cellcolor{blue!10}63.41 & \cellcolor{blue!10}82.95 & \cellcolor{blue!10}\textbf{92.96} & \cellcolor{blue!10}\textbf{82.47} & \cellcolor{blue!10}85.18 & \cellcolor{blue!10}70.99 \\
 &  & \cellcolor{red!10}\textbf{GEM} & \cellcolor{red!10}90.48 & \cellcolor{red!10}45.76 & \cellcolor{red!10}\underline{\emph{90.48}} & \cellcolor{red!10}\underline{\emph{80.30}} & \cellcolor{red!10}94.59 & \cellcolor{red!10}78.10 & \cellcolor{red!10}87.80 & \cellcolor{red!10}92.06 & \cellcolor{red!10}70.00 & \cellcolor{red!10}70.43 & \cellcolor{red!10}84.04 & \cellcolor{red!10}\underline{\emph{78.29}} \\
 &  & \cellcolor{red!10}\textbf{\model} & \cellcolor{red!10}\textbf{97.62} & \cellcolor{red!10}\textbf{55.95} & \cellcolor{red!10}\textbf{93.81} & \cellcolor{red!10}\textbf{86.30} & \cellcolor{red!10}\textbf{100.00} & \cellcolor{red!10}\textbf{94.42} & \cellcolor{red!10}87.80 & \cellcolor{red!10}88.48 & \cellcolor{red!10}77.78 & \cellcolor{red!10}70.14 & \cellcolor{red!10}\textbf{88.54} & \cellcolor{red!10}\textbf{83.78} \\
 
\bottomrule
\end{tabular}

\vspace{-15pt}
\label{tab:alignment}
\end{table*}

\subsection{LLM Backbone and Training Objective}
\label{sec:backbone}

The enhanced numerical and visual token embeddings $\{\hat{\mathbf{E}}^n_i,\hat{\mathbf{E}}^v_i\}_{i=1}^{|\mathcal{X}|}$ are concatenated with textual input tokens, including the question embeddings $\mathbf{E}^q$ and context embeddings $\mathbf{E}^c$ (derived from the context $C$ using the LLM's embeddings), via a position-aware concatenation operator~\cite{chatts2025}:
\begin{equation}
\mathbf{E}^*=\text{PosConcat}(\mathbf{E}^c,\mathbf{E}^q,\{\hat{\mathbf{E}}^n_i,\hat{\mathbf{E}}^v_i\}_{i=1}^{|\mathcal{X}|}),
\end{equation}
where $\text{PosConcat}(\cdot)$ integrates multi-modal token embeddings in the textual inputs according to the positions of the time series instances in the original input.
The fused sequence $\mathbf{E}^*$ is fed into the backbone LLM to generate the answer $A$ autoregressively: $P(A \mid \mathbf{E}^*) = \prod_{b=1}^{|A|} P(A_b \mid A_{<b}, \mathbf{E}^*)$, where $A_b$ denotes the $b$-th token of the target answer sequence.
In this study, we adopt Qwen2.5-VL-7B-Instruct~\cite{qwen25vl} as our backbone MLLM.
For model training, the language modeling loss is:
$\mathcal{L}_{\text{LM}}=-\sum_{b=1}^{|A|}\log P(A_b\mid A_{<b},\mathbf{E}^*)$, 
and the overall training objective incorporates the regularization terms from PA and DDI: $\mathcal{L}=\mathcal{L}_{\text{LM}}+\lambda_1\cdot\mathcal{L}_{\text{PA}}+\lambda_2\cdot\mathcal{L}_{\text{DDI}}$, where $\lambda_1,\lambda_2$ denote the loss weights.

\section{Experiment}
\label{sec:exp}

\subsection{Experimental Setups}

\textbf{Dataset.} We use the training and evaluation data released with ChatTS~\cite{chatts2025}, consisting of synthetic language–time series pairs for training and a hybrid of synthetic and real-world set for evaluation. 
The evaluation covers two types of tasks: 
(1) \emph{Understanding tasks} assess fundamental characteristics of uni- or multi-variate time series, including noise, local fluctuation, seasonality, trend, as well as correlation and clustering patterns across variables; 
(2) \emph{Reasoning tasks} involve inductive, deductive, causal, and comparison reasoning questions. 
Understanding tasks may require categorical choices or numerical calculations, whereas reasoning tasks mainly involve categorical questions, with some open-ended questions for inductive reasoning. 
Detailed descriptions of the datasets are provided in \appref{app:dataset}.

\textbf{Baseline.} We compare our model against three types of baselines, grouped by the formulation of time series modalities:
(1) \emph{Numerical-centric models} directly process time series values, including general LLMs (GPT-4o~\cite{gpt4o}, Qwen3~\cite{qwen3}, DeepSeek-V3.2~\cite{deepseekv32}, Gemini 3 Pro~\cite{gemini3pro}, GPT-5.2~\cite{gpt52}) which formulate numerical values into textual prompts, and time series-specific MLLMs (ChatTime~\cite{chattime2025}, ChatTS~\cite{chatts2025}, ITFormer~\cite{itformer2025}, InstructTime~\cite{instructtime2025}) which introduce specialized time series encoders; 
(2) \emph{Visual-centric models} leverage time series plots (GPT-4o, Qwen3-VL~\cite{qwen3vl}, Gemini 3 Pro, GPT-5.2); and 
(3) \emph{Numerical+Visual models} integrate both modalities (GPT-4o, Qwen3-VL, Gemini 3 Pro, GPT-5.2, GEM~\cite{gem2025}). 
General-purpose LLMs/MLLMs are accessed via APIs, while time-series-specialized MLLMs are fine-tuned on unified LLM backbones and training data. 
More details are provided in \appref{app:baseline}.

\textbf{Evaluation Metric.} For categorical questions, we report \emph{Accuracy} or \emph{F1-Score}; for numerical questions, we use \emph{Relative Accuracy}~\cite{chatts2025}. 
For open-ended reasoning questions, we adopt \emph{Answer Correctness}~\cite{ragas2024}, which measures the extent to which the model response covers the minimal set of essential facts (extracted by GPT-4o-mini~\cite{gpt4omini}) required to answer the question. 

The implementation details are provided in \appref{app:implement}.

\subsection{Comparison Results}

\tabref{tab:alignment} reports performance on understanding tasks across synthetic and real-world datasets, while \tabref{tab:reasoning} summarizes results of reasoning tasks on the hybrid dataset. 
From these results, we draw the following key observations::
(1) \emph{\model\xspace demonstrates strong and consistent understanding/reasoning capabilities.} 
It achieves clear improvements across both understanding and reasoning tasks, indicating that carefully designed multi-modal alignment and interaction mechanisms are crucial for effective TSUR.
(2) \emph{Visual plots are generally more effective than numerical sequences.} 
Visual-centric models outperform numerical-centric counterparts in most settings. 
For example, GPT-4o achieves up to 17\% higher categorical accuracy on real-world understanding tasks.
(3) \emph{Naively combining modalities is insufficient.} 
For general-purpose LLMs, multi-modal inputs slightly degrade performance relative to visual-centric models, revealing the deficiency of simple modality concatenation, and the need for structured, fine-grained multi-modal learning strategies.
(4) \emph{General-purpose LLMs generalize better to real-world scenarios.} 
While their performance on synthetic datasets is modest (categorical accuracy $\sim$66\%), general-purpose LLMs achieve substantially higher scores on real-world time series. 
For example, visually augmented Gemini 3 Pro outperforms several time-series-specialized MLLMs, such as ChatTS and GEM, on categorical understanding. 
In contrast, models fine-tuned on synthetic data often degrade on real-world benchmarks, particularly in queries about cross-variable correlation and clustering. 
While this gap is partially attributable to model scale, it also reflects the challenge of robust real-world generalization in TSUR.
(5) \emph{Reasoning tasks are significantly harder than understanding tasks.} 
Performance on reasoning tasks is substantially lower than that on understanding tasks, emphasizing the intrinsic difficulty of higher-level time series reasoning and motivating further research in this direction.
In addition, a computational cost analysis is provided in \appref{app:token_cost}, and detailed case studies are presented in \appref{app:case}.

\begin{table}[t]
\centering
\caption{Performance (\%) on reasoning tasks.}
\vspace{-5pt}
\resizebox{\linewidth}{!}{
\scriptsize
\begin{tabular}{c | c | c c c c c}
\toprule
\textbf{Type} & \textbf{Model} & \textbf{Induct.}& \textbf{Deduct.}& \textbf{Causal}& \textbf{MCQ2}& \textbf{Overall} \\
\midrule
\multirow{9}{*}{\rotatebox{90}{\textbf{Numerical}}} & \cellcolor{blue!10}\textbf{DeepSeek-V3.2} & \cellcolor{blue!10}33.66 & \cellcolor{blue!10}\textbf{69.77} & \cellcolor{blue!10}72.83 & \cellcolor{blue!10}56.00 & \cellcolor{blue!10}40.99 \\
 & \cellcolor{blue!10}\textbf{GPT-4o} & \cellcolor{blue!10}32.07 & \cellcolor{blue!10}60.47 & \cellcolor{blue!10}66.30 & \cellcolor{blue!10}49.00 & \cellcolor{blue!10}38.22 \\
 & \cellcolor{blue!10}\textbf{Qwen3} & \cellcolor{blue!10}32.38 & \cellcolor{blue!10}48.84 & \cellcolor{blue!10}68.48 & \cellcolor{blue!10}46.00 & \cellcolor{blue!10}37.93 \\
 & \cellcolor{blue!10}\textbf{GPT-5.2} & \cellcolor{blue!10}30.32 & \cellcolor{blue!10}0.00 & \cellcolor{blue!10}73.91 & \cellcolor{blue!10}43.00 & \cellcolor{blue!10}34.23 \\
 & \cellcolor{blue!10}\textbf{Gemini 3 Pro} & \cellcolor{blue!10}34.46 & \cellcolor{blue!10}46.51 & \cellcolor{blue!10}63.04 & \cellcolor{blue!10}\underline{\emph{68.00}} & \cellcolor{blue!10}40.90 \\
 & \cellcolor{red!10}\textbf{ChatTime} & \cellcolor{red!10}49.18 & \cellcolor{red!10}48.84 & \cellcolor{red!10}53.26 & \cellcolor{red!10}44.00 & \cellcolor{red!10}49.07 \\
 & \cellcolor{red!10}\textbf{ChatTS} & \cellcolor{red!10}55.66 & \cellcolor{red!10}51.16 & \cellcolor{red!10}58.70 & \cellcolor{red!10}53.00 & \cellcolor{red!10}55.50 \\
 & \cellcolor{red!10}\textbf{ITFormer} & \cellcolor{red!10}42.54 & \cellcolor{red!10}44.19 & \cellcolor{red!10}47.83 & \cellcolor{red!10}46.00 & \cellcolor{red!10}43.48 \\
 & \cellcolor{red!10}\textbf{InstructTime} & \cellcolor{red!10}47.02 & \cellcolor{red!10}53.49 & \cellcolor{red!10}50.00 & \cellcolor{red!10}52.00 & \cellcolor{red!10}48.25 \\
\midrule
\multirow{4}{*}{\rotatebox{90}{\textbf{Visual}}} & \cellcolor{blue!10}\textbf{GPT-4o} & \cellcolor{blue!10}28.86 & \cellcolor{blue!10}62.79 & \cellcolor{blue!10}68.48 & \cellcolor{blue!10}51.00 & \cellcolor{blue!10}36.03 \\
 & \cellcolor{blue!10}\textbf{Qwen3-VL} & \cellcolor{blue!10}33.11 & \cellcolor{blue!10}51.16 & \cellcolor{blue!10}60.87 & \cellcolor{blue!10}55.00 & \cellcolor{blue!10}38.75 \\
 & \cellcolor{blue!10}\textbf{GPT-5.2} & \cellcolor{blue!10}32.09 & \cellcolor{blue!10}51.16 & \cellcolor{blue!10}67.39 & \cellcolor{blue!10}51.00 & \cellcolor{blue!10}37.93 \\
 & \cellcolor{blue!10}\textbf{Gemini 3 Pro} & \cellcolor{blue!10}36.57 & \cellcolor{blue!10}62.79 & \cellcolor{blue!10}70.65 & \cellcolor{blue!10}\underline{\emph{68.00}} & \cellcolor{blue!10}43.87 \\
\midrule
\multirow{6}{*}{\rotatebox{90}{\textbf{Num.+Visual}}} & \cellcolor{blue!10}\textbf{GPT-4o} & \cellcolor{blue!10}26.81 & \cellcolor{blue!10}55.81 & \cellcolor{blue!10}69.57 & \cellcolor{blue!10}47.00 & \cellcolor{blue!10}33.93 \\
 & \cellcolor{blue!10}\textbf{Qwen3-VL} & \cellcolor{blue!10}29.70 & \cellcolor{blue!10}51.16 & \cellcolor{blue!10}69.57 & \cellcolor{blue!10}54.00 & \cellcolor{blue!10}36.76 \\
 & \cellcolor{blue!10}\textbf{GPT-5.2} & \cellcolor{blue!10}29.93 & \cellcolor{blue!10}34.88 & \cellcolor{blue!10}\underline{\emph{75.00}} & \cellcolor{blue!10}46.00 & \cellcolor{blue!10}35.74 \\
 & \cellcolor{blue!10}\textbf{Gemini 3 Pro} & \cellcolor{blue!10}30.12 & \cellcolor{blue!10}53.49 & \cellcolor{blue!10}58.70 & \cellcolor{blue!10}\textbf{69.00} & \cellcolor{blue!10}37.49 \\
 & \cellcolor{red!10}\textbf{GEM} & \cellcolor{red!10}\textbf{58.73} & \cellcolor{red!10}\underline{\emph{67.44}} & \cellcolor{red!10}66.30 & \cellcolor{red!10}53.00 & \cellcolor{red!10}\underline{\emph{59.30}} \\
 & \cellcolor{red!10}\textbf{\model} & \cellcolor{red!10}\underline{\emph{57.68}} & \cellcolor{red!10}\underline{\emph{67.44}} & \cellcolor{red!10}\textbf{76.09} & \cellcolor{red!10}62.00 & \cellcolor{red!10}\textbf{61.98} \\
\bottomrule
\end{tabular}

}
\vspace{-15pt}
\label{tab:reasoning}
\end{table}

\subsection{Ablation Study}

\begin{figure}[h]
    \centering
    \vspace{-5pt}
    \includegraphics[width=\linewidth]{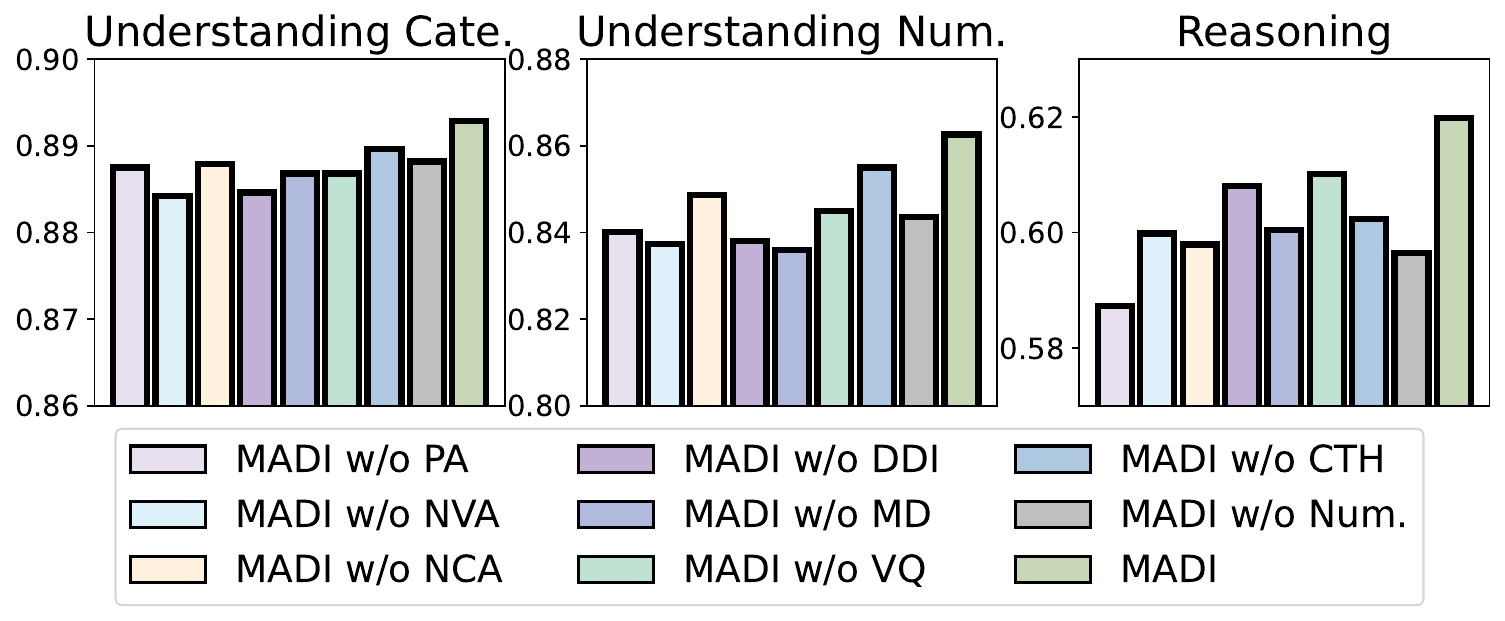}
    \vspace{-18pt}
    \caption{Ablation results of \model.}
    \vspace{-10pt}
    \label{fig:ablation}
\end{figure}

We assess the contribution of individual components by comparing \model\xspace with several ablated variants. “w/o PA”, “w/o DDI”, and “w/o CTH” remove the PA, DDI, and CTH modules, respectively. To evaluate fine-grained alignment and disentanglement, “w/o NVA” removes numerical-visual alignment, “w/o NCA” removes numerical-caption alignment, and “w/o MD” eliminates modality disentanglement. Furthermore, “w/o VQ” replaces hierarchical discretization in DDI with modality-shared continuous encoders, and “w/o Num.” removes the numerical modality, disabling PA and DDI.
\figref{fig:ablation} reports performance on categorical and numerical understanding, as well as reasoning tasks, revealing: 
(1) Alignment and interaction between different modalities are essential, as removing PA or DDI leads to substantial performance drops. 
(2) Highlighting critical, question-relevant information improves robustness and generalization, particularly for unseen reasoning tasks (“w/o CTH”). 
(3) Both numerical-visual and numerical-caption alignments are critical, as the performance gap between “w/o NVA”, “w/o NCA”, and the full model indicates a synergistic effect when both are employed. 
(4) Modality disentanglement (“w/o MD”) plays a pivotal role, as isolated unique signals significantly facilitate cross-modal interaction. 
(5) VQ is crucial for effective modality disentanglement and interaction, shown by the decrease in “w/o VQ”. 
(6) Numerical information contributes to robustness and generalization, but its benefits depend on well-designed multi-modal alignment and interaction (“w/o Num.” vs. full model).

\subsection{Visualized Analysis}

In this section, we empirically evaluate \model’s ability to address the two key challenges identified in \secref{sec:intro}.

\begin{figure}[h]
    \centering
    \vspace{-10pt}
    \includegraphics[width=\linewidth]{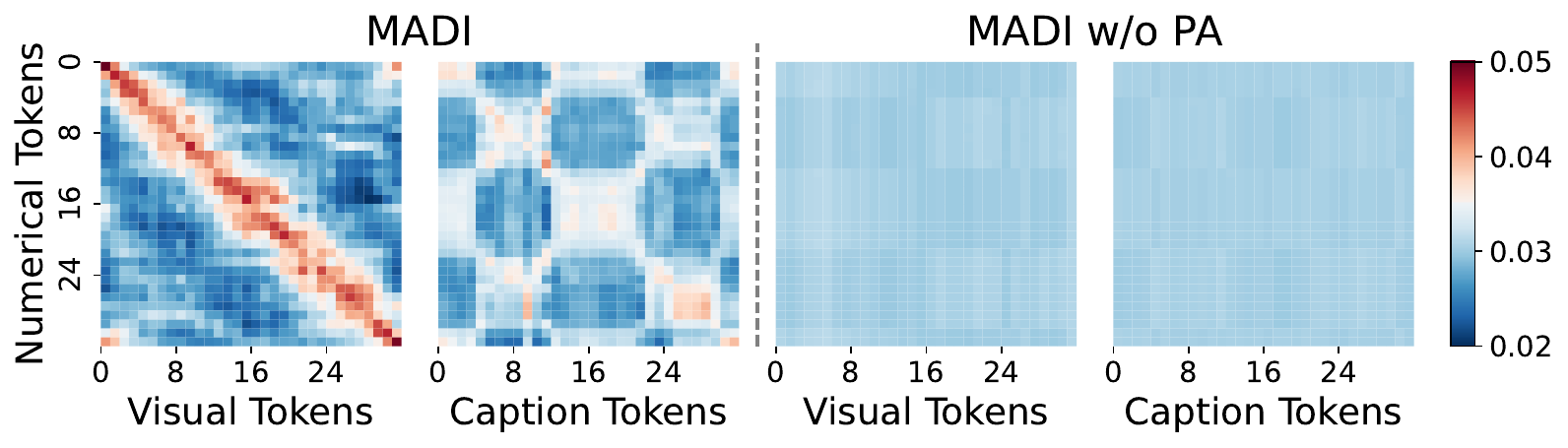}
    \vspace{-15pt}
    \caption{Heatmaps of cross-modal patch embedding similarities.}
    \vspace{-8pt}
    \label{fig:align}
\end{figure}

\textbf{Alignment Analysis.} 
To verify the effectiveness of fine-grained cross-modal grounding, we visualize the patch-wise cosine similarity matrices between numerical embeddings and their visual/textual counterparts in \figref{fig:align}. 
Compared to the ablated variant (\emph{w/o PA}), \model\xspace exhibits clear diagonal high-similarity patterns, particularly between numerical and visual tokens. 
This confirms that the PA module successfully enforces precise patch-level correspondence across modalities. 
We further observe that the numerical-caption alignment is relatively less pronounced than the numerical-visual one, attributable to the semantic redundancy introduced by template-based caption generation, which makes patch-to-text consistency inherently more ambiguous.

\begin{figure}[h]
    \centering
    \vspace{-5pt}
    \includegraphics[width=\linewidth]{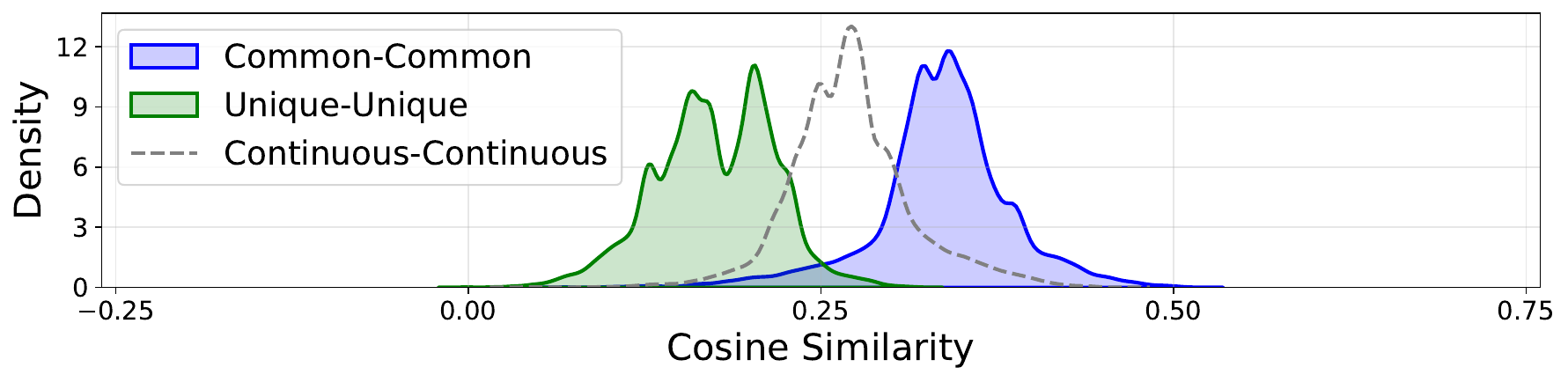}
    \vspace{-15pt}
    \caption{Distributions of cross-modal embedding similarities.}
    \vspace{-15pt}
    \label{fig:disentangle}
\end{figure}

\textbf{Disentanglement Analysis.} 
To assess the quality of semantic decoupling in the DDI module, we analyze the kernel density estimation (KDE) distribution of cross-modal cosine similarities in \figref{fig:disentangle}. 
We compare the similarity distributions of original continuous embeddings (\emph{Continuous-Continuous}), disentangled modality-common signals (\emph{Common-Common}), and modality-unique residuals (\emph{Unique-Unique}). 
The results show that modality-common pairs exhibit higher similarity than their continuous counterparts, indicating successful concentration of shared semantics.
In contrast, modality-unique pairs display substantially reduced correlation, verifying that DDI effectively isolates unique information from the shared subspace.

\section{Conclusion}
\label{sec:conclusion}

We present \model, a multi-modal LLM for TSUR tasks that addresses the challenges of fine-grained alignment and disentangled interaction between numerical and visual modalities.
The PA module enforces physically grounded correspondence across numerical, visual, and textual tokens, while the DDI module separates modality-common and modality-specific semantics to synergistically fuse complementary signals.
The CTH module further emphasizes informative, query-relevant tokens to enhance reasoning robustness.
Extensive experiments on synthetic and real-world benchmarks demonstrate that \model\xspace consistently outperforms general-purpose LLMs and time-series-specialized MLLMs, validating the effectiveness of structured multi-modal integration and highlighting its potential to advance interpretable and flexible time series analysis.

\section*{Impact Statement}

This work aims to advance the field of machine learning by developing a numerical-visual MLLM, \model, for interpretable and flexible time series understanding and reasoning. 
By effectively integrating numerical precision and visual abstraction, \model\xspace enhances the ability of AI systems to analyze temporal data across diverse domains such as healthcare, finance, climate, and infrastructure monitoring. 
While the model is designed for general-purpose time series analysis, we recognize that its deployment in sensitive real-world applications, such as medical diagnosis, financial forecasting, or autonomous systems, requires careful consideration of data privacy, algorithmic fairness, and robustness to distribution shifts. 
We encourage future work to incorporate ethical safeguards, domain-specific validation, and human-in-the-loop oversight to mitigate potential risks associated with misinterpretation or over-reliance on automated decisions. 
Ultimately, we believe this research contributes to the development of more transparent, adaptable, and trustworthy AI tools for time-series analytics, with broadly beneficial societal implications.

\nocite{langley00}

\bibliography{ref_formal}
\bibliographystyle{icml2026}

\newpage
\appendix
\onecolumn
\section{Notation}
\label{app:notation}

The key notations used in this paper are summarized in Table~\ref{table:notation}.

\begin{table*}[h!]
\centering
\caption{Summary of Notations.}
\resizebox{0.75\linewidth}{!}{
\begin{tabular}{ll}
\toprule
\textbf{Notation} & \textbf{Description} \\
\midrule
\multicolumn{2}{l}{\textbf{General \& Problem Definition}} \\
$\mathcal{X}$ & Set of time series inputs, containing instances $\mathbf{x}_i$ \\
$\mathbf{x}_i$ & The $i$-th time series instance \\
$T_i$ & Number of observations in time series $\mathbf{x}_i$ \\
$C, Q$ & Context and question textual inputs \\
$\mathbf{E}^c, \mathbf{E}^q$ & Embeddings for context and question, composed of tokens $\mathbf{e}^c, \mathbf{e}^q$ \\
$A$ & Generated textual answer \\
\midrule
\multicolumn{2}{l}{\textbf{Patch-level Alignment (PA)}} \\
$p^n$ & Patch size for numerical time series \\
$p^v$ & Patch size for the vision encoder \\
$\tilde{T}_i$ & Number of patches for $\mathbf{x}_i$ \\
$\mathbf{P}^n_i$ & Numerical time series patches, composed of $\mathbf{p}^n_{i;j}$ \\
$\mathbf{I}_i$ & Visualized line plot image \\
$S_i$ & Set of patch-wise textual captions $\{s_{i;j}\}$ \\
$D$ & Embedding dimension of above encoders and the LLM backbone \\
$\mathbf{E}^n_i$ & Encoded numerical embeddings, composed of tokens $\mathbf{e}^n_{i;j}$ \\
$\mathbf{E}^v_i$ & Encoded visual embeddings, composed of tokens $\mathbf{e}^v_{i;j}$ \\
$\mathbf{E}^s_i$ & Encoded caption embeddings, composed of tokens $\mathbf{e}^s_{i;j}$
\\
\midrule
\multicolumn{2}{l}{\textbf{Discrete Disentangled Interaction (DDI)}} \\
$\mathbf{Z}^n_i, \mathbf{Z}^v_i$ & Modality-common representations, composed of tokens $\mathbf{z}^n_{i;j}, \mathbf{z}^v_{i;j}$ \\
$\mathbf{U}^n_i, \mathbf{U}^v_i$ & Modality-unique representations, composed of tokens $\mathbf{u}^n_{i;j}, \mathbf{u}^v_{i;j}$ \\
$d$ & Reduced embedding dimension during vector quantization \\
$\mathcal{C}^{(m)}$ & The $m$-th codebook in hierarchical vector quantization \\
$\mathbf{R}^{n;(m)}_i, \mathbf{R}^{v;(m)}_i$ & Residuals at quantization stage $m$, composed of tokens $\mathbf{r}^{n;(m)}_{i;j}, \mathbf{r}^{v;(m)}_{i;j}$ \\
$\tilde{\mathbf{R}}^{n;(m)}_i, \tilde{\mathbf{R}}^{v;(m)}_i$ & Pooled and broadcasted residuals at stage $m$, composed of tokens $\tilde{\mathbf{r}}^{n;(m)}_{i;j}, \tilde{\mathbf{r}}^{v;(m)}_{i;j}$ \\
$\mathbf{Q}^{n;(m)}_i, \mathbf{Q}^{v;(m)}_i$ & Quantized discrete modality representations, composed of $\mathbf{q}^{n;(m)}_{i;j}, \mathbf{q}^{v;(m)}_{i;j}$ \\
$\overline{\mathbf{E}}^n_i, \overline{\mathbf{E}}^v_i$ & Augmented modality tensors after cross-modal interaction \\
\midrule
\multicolumn{2}{l}{\textbf{Critical-token Highlighting (CTH)}} \\
$\mathbf{Q}^q$ & Learnable queries for the question-conditioned branch \\
$\mathbf{Q}^n, \mathbf{Q}^v$ & Learnable queries for the modality-intrinsic branch \\
$\mathbf{H}^{n;q}_i, \mathbf{H}^{v;q}_i$ & Highlighted tokens from the question-conditioned branch \\
$\mathbf{H}^{n;s}_i, \mathbf{H}^{v;s}_i$ & Highlighted tokens from the modality-intrinsic branch \\
$\mathbf{H}^n_i, \mathbf{H}^v_i$ & Highlighted tokens fused from question-conditioned and modality-intrinsic branches \\
$\hat{\mathbf{E}}^n_i, \hat{\mathbf{E}}^v_i$ & Final enhanced modality embeddings prepended with highlighted tokens \\
$\mathbf{E}^*$ & Final fused multi-modal sequence input to the LLM \\
\midrule
\multicolumn{2}{l}{\textbf{Optimization Objectives \& Hyperparameters}} \\
$\mathcal{L}_{PA}$ & Total Patch-level Alignment loss \\
$\mathcal{L}_{align}^{n\text{-}v}, \mathcal{L}_{align}^{n\text{-}s}$ & Contrastive alignment losses for numerical-visual and numerical-caption pairs \\
$\mathcal{L}_{DDI}$ & Total Discrete Disentangled Interaction loss \\
$\mathcal{L}_{vq}$ & Vector quantization commitment loss \\
$\mathcal{L}_{com}$ & Bidirectional contrastive loss for modality-common semantics \\
$\mathcal{L}_{orth}$ & Orthogonality loss between common and unique components \\
$\mathcal{L}_{LM}$ & Language modeling loss \\
$\tau$ & Temperature hyperparameter for contrastive learning \\
$\alpha, \beta$ & Weights for disentanglement regularization terms \\
$\lambda_1, \lambda_2$ & Loss weights for PA and DDI modules \\
\bottomrule
\end{tabular}
}
\label{table:notation}
\end{table*}

\section{Experimental Details}

\subsection{Dataset}
\label{app:dataset}

Our experiments utilize the datasets released by ChatTS~\cite{chatts2025}, which are designed for time series understanding and reasoning (TSUR). 
The data are divided into training and evaluation sets and are sourced from both synthetic generation and real-world domains.

\subsubsection{Training Data Construction}
\label{app:dataset_train}

Due to the scarcity of high-quality, well-aligned time series--text pairs, the training data are primarily constructed via a synthetic generation pipeline~\cite{chatts2025} consisting of two key modules:
\begin{itemize}
    \item \textbf{Attribute-Based Generation:} To ensure precise alignment between numerical values and textual descriptions, time series are generated from a predefined \emph{Attribute Pool} covering four major categories: \emph{Trend}, \emph{Periodicity}, \emph{Noise}, and \emph{Local Fluctuations} (e.g., spikes and phase changes). 
    An \emph{Attribute Selector} samples attributes to construct time series arrays while simultaneously generating ground-truth textual descriptions based on the selected attributes.
    \item \textbf{Time Series Evol-Instruct (TSEvol):} To enhance data diversity and reasoning complexity, we adopt an evolutionary instruction generation strategy. Starting from seed question--answer pairs, an LLM progressively evolves questions into more complex forms, while querying the attribute pool to ensure factual consistency.
\end{itemize}

\subsubsection{Evaluation Data Composition}

The evaluation dataset is a hybrid of real-world and synthetic data, designed to assess out-of-distribution generalization. 
The synthetic samples are generated using the same pipeline described in~\appref{app:dataset_train}. 
The real-world data are collected from diverse domains, including AIOps (system monitoring metrics), weather (temperature and humidity), finance (stock trends), and traffic flow. 
All real-world samples are manually annotated by domain experts to ensure annotation accuracy.

Concretely, the evaluation dataset comprises two primary task categories:
\begin{itemize}
    \item \textbf{Understanding Tasks:} These tasks evaluate the model's ability to ground textual concepts in time series features. Subtasks include identifying \emph{Trend} (increase/decrease), \emph{Seasonality} (period length), \emph{Noise Levels}, and \emph{Local Fluctuations} (spike/dip detection). 
    For multivariate time series, additional tasks include \emph{Correlation} analysis and \emph{Clustering}. 
    Questions are presented in both categorical (multiple-choice) and numerical (value extraction) formats.
    \item \textbf{Reasoning Tasks:} These tasks assess higher-order reasoning capabilities. Subtasks include \emph{Inductive Reasoning} (inferring physical meanings from patterns), \emph{Deductive Reasoning} (predicting behavior under given conditions), \emph{Causal Reasoning} (identifying root causes of anomalies), and \emph{Comparison} (contrasting distinct time series). 
    Questions are provided in categorical or open-ended formats.
\end{itemize}

\subsection{Baselines}
\label{app:baseline}

To ensure rigorous comparisons, we group baselines into general-purpose LLMs/MLLMs and specialized time series MLLMs.

\textbf{General LLMs and MLLMs.}
We evaluate general-purpose models including GPT-4o~\cite{gpt4o}, Qwen3~\cite{qwen3}, Qwen3-VL~\cite{qwen3vl}, DeepSeek-V3.2~\cite{deepseekv32}, Gemini 3 Pro~\cite{gemini3pro}, and GPT-5.2~\cite{gpt52}, using their official APIs. 
Numerical-only LLMs receive time series in serialized numerical form, while visual-based MLLMs additionally receive time series rendered as images.
Following~\cite{chatts2025}, we visualize time series using line plots; for multivariate data, plots of different variables are stacked into a single composite figure.

\textbf{Time-Series-Specialized MLLMs.}
For specialized baselines, including ChatTime~\cite{chattime2025}, ChatTS~\cite{chatts2025}, ITFormer~\cite{itformer2025}, InstructTime~\cite{instructtime2025}, and GEM~\cite{gem2025}, we unify the LLM backbone to ensure a fair comparison that emphasizes architectural design rather than base model capacity. 
All specialized models are reproduced and fine-tuned using either Qwen2.5-7B-Instruct~\cite{qwen25} or Qwen2.5-VL-7B-Instruct~\cite{qwen25vl} for numerical--visual multimodal settings. 
We strictly maintain identical training data and experimental environments across all baselines and our proposed \model.

\subsection{Evaluation Metrics}
\label{app:metric}

We adopt different evaluation metrics for categorical, numerical, and open-ended reasoning tasks.

\textbf{Categorical Tasks.}
For classification-based questions (e.g., trend direction and fluctuation type), we report \emph{Accuracy} and \emph{F1-score}, measuring the agreement between predicted labels and ground truth.

\textbf{Numerical Tasks.}
For tasks requiring numerical value estimation (e.g., spike amplitude or period length), we employ \emph{Relative Accuracy} ($\text{Acc}_{\text{rel}}$) with thresholding to reduce sensitivity to outliers:
\begin{equation}
\text{Acc}_{\text{rel}} = \max \left( 0, 1 - \left| \frac{v_{\text{pred}} - v_{\text{label}}}{v_{\text{label}}} \right| \right),
\end{equation}
where $v_{\text{pred}}$ denotes the value extracted from the model response and $v_{\text{label}}$ is the ground-truth value.

\textbf{Open-ended Reasoning Tasks.}
For open-ended reasoning tasks, we adopt the \emph{Answer Correctness} metric from RAGAS framework~\cite{ragas2024}. 
This metric uses a judge LLM (GPT-4o-mini) to extract key facts from the ground-truth answer and computes the recall of these facts in the model response, enabling semantic evaluation beyond naive keyword matching.

\subsection{Implementation Details}
\label{app:implement}

This section describes the implementation details of \model, including data preprocessing, model hyperparameters, and training and evaluation configurations.

\textbf{Statistics-Preserved Normalization.}
To improve training stability and reduce distribution shifts across heterogeneous time series domains, we apply zero-centered normalization to the input series prior to patch-level modality expansion.
While normalization is essential for effective optimization, it removes absolute magnitude information that is crucial for precise numerical reasoning (e.g., answering queries such as ``what is the maximum value?'').

To address this limitation, we introduce a \textit{Statistics-Preserved Prompt} that explicitly encodes global statistics of the original time series and exposes them to the LLM during reasoning.
Specifically, for each time series instance $\mathbf{x}_i$, we construct a structured textual prompt containing its normalization parameters (offset and scaling factor), sequence length, and key boundary statistics, including the maximum, minimum, start, and end values.
This prompt is prepended to the processed multi-modal tokens $\hat{\mathbf{E}}^n_i$, enabling the LLM to recover the original scale and perform accurate value-based reasoning.

The statistics-preserved prompt follows the format:
\begin{equation}
    \texttt{[offset=...} \texttt{|scaling=...} \texttt{|length=...} \texttt{|max=...} \texttt{|min=...} \texttt{|left=...} \texttt{|right=...]}
\end{equation}
where:
\begin{itemize}
    \item \texttt{offset}: the offset applied during normalization (\eg the negative mean);
    \item \texttt{scaling}: the scaling factor used for normalization;
    \item \texttt{length}: the length of the time series sequence;
    \item \texttt{max} / \texttt{min}: the maximum and minimum values of the original series;
    \item \texttt{left} / \texttt{right}: the first and last values in the sequence.
\end{itemize}

By exposing these global statistics, \model\xspace leverages the arithmetic reasoning capabilities of the LLM backbone to interpret normalized temporal patterns with respect to their original physical magnitudes.
For consistency across modalities, visual plots and patch-level captions are also generated from the normalized series, ensuring alignment with the numerical inputs.

\textbf{Model Hyperparameters.}
The embedding dimension is set to $D = 3584$ to match the LLM backbone. 
We use a numerical patch size of $p^n = 8$, a discretization dimension of $d = 512$, and $m = 3$ hierarchical codebook levels. 
The overall loss is a weighted sum of multiple components, with weights $\alpha = 5$ (modality-common alignment), $\beta = 1$ (common--unique orthogonalization), $\lambda_1 = 0.02$ (PA regularization), and $\lambda_2 = 0.2$ (DDI regularization).

\textbf{Training Configuration.}
We implement our model and training pipeline using the LLaMA-Factory framework on $4 \times$ NVIDIA A800 GPUs.
Specifically, we adopt a per-device batch size of 2 with 64 gradient accumulation steps, resulting in an effective batch size of $512 = 4 \times 2 \times 64$. 
Optimization is performed using AdamW with a learning rate of $1 \times 10^{-5}$ and a cosine learning rate scheduler, together with a warmup ratio of 0.02. 
Training proceeds for 1200 steps.
We employ a two-stage training strategy: during the initial warmup phase (0.02 of total training), the LLM backbone and vision encoder (for Qwen2.5-VL-based models) are frozen, and only newly introduced modules are trained; afterward, all parameters are unfrozen for full fine-tuning.

\textbf{Evaluation Configuration.}
During evaluation, we set the decoding temperature to 0.01 to ensure stable and reproducible results across all benchmarks.
Evaluation is conducted every 50 steps, and the checkpoint with the best evaluation performance is selected for final reporting.

\section{Token Cost Analysis}
\label{app:token_cost}

In this section, we evaluate the computational efficiency of different paradigms by calculating the average token consumption per query in real-world understanding tasks, as presented in Table \ref{tab:token_cost}. 
This reveals significant disparities driven by tokenization strategies and input modalities.

\begin{table}[h]
\centering
\caption{Average token cost per query across different kinds of models.}
\label{tab:token_cost}

\resizebox{0.9\linewidth}{!}{
\begin{tabular}{l|ccccc|cccc}
\toprule
\textbf{Type} & \multicolumn{9}{c}{\textbf{Numerical}} \\
\cmidrule{1-10}
\textbf{Model} & 
\textbf{DeepSeek} & \textbf{GPT-4o} & \textbf{Qwen3} & \textbf{GPT-5.2} & \textbf{Gemini} & 
\textbf{InstructTime} & \textbf{ITFormer} & \textbf{ChatTime} & \textbf{ChatTS} \\
\midrule
\textbf{Cost} & 
\cellcolor{blue!10}7135.90 & \cellcolor{blue!10}7140.04 & \cellcolor{blue!10}9935.51 & \cellcolor{blue!10}7139.04 & \cellcolor{blue!10}9702.75 & 
\cellcolor{red!10}1108.16 & \cellcolor{red!10}1108.16 & \cellcolor{red!10}2503.01 & \cellcolor{red!10}1118.90 \\
\bottomrule
\end{tabular}
}

\vspace{2pt} 

\resizebox{0.9\linewidth}{!}{
\begin{tabular}{l|cccc|cccccc}
\toprule
\textbf{Type} & \multicolumn{4}{c|}{\textbf{Visual}} & \multicolumn{6}{c}{\textbf{Num.+Visual}} \\
\cmidrule{1-11}
\textbf{Model} & 
\textbf{GPT-4o} & \textbf{Qwen3-VL} & \textbf{GPT-5.2} & \textbf{Gemini} & 
\textbf{GPT-4o} & \textbf{Qwen3-VL} & \textbf{GPT-5.2} & \textbf{Gemini} & \textbf{GEM} & \textbf{\model} \\
\midrule
\textbf{Cost} & 
\cellcolor{blue!10}1063.49 & \cellcolor{blue!10}1390.13 & \cellcolor{blue!10}1111.86 & \cellcolor{blue!10}1643.47 & 
\cellcolor{blue!10}7698.82 & \cellcolor{blue!10}10800.71 & \cellcolor{blue!10}7864.87 & \cellcolor{blue!10}10800.93 & \cellcolor{red!10}1174.99 & \cellcolor{red!10}1517.79 \\
\bottomrule
\end{tabular}
}
\end{table}

In the \textit{Numerical-centric} setting, general-purpose LLMs (\eg GPT-4o, DeepSeek) exhibit prohibitively high consumption, consistently exceeding 7,000 tokens per query. This inefficiency stems from their reliance on digit-level tokenization, where a single time-series value is decomposed into multiple tokens. 
Even time-series-specific adaptations like ChatTime, which utilizes value-level tokenization, incur a noticeably higher cost compared to patch-based models (\eg ITFormer, ChatTS, \model) that efficiently aggregate temporal segments into compact vector representations.

The computational burden is further exacerbated in the \textit{Numerical+Visual} setting. 
For standard Multi-modal LLMs such as Qwen3-VL and GPT-4o, the straightforward concatenation of lengthy numerical text sequences with visual embeddings results in excessive sequence lengths, often surpassing 10,000 tokens. 
This significantly increases the inference overhead. 
In contrast, \model\xspace incorporates rich multi-modal semantics with only a marginal increase in token count compared to the most efficient baselines. 
Consequently, \model\xspace achieves the best performance trade-off, delivering robust reasoning capabilities with a modest token cost.

\section{Case Study}
\label{app:case}

In this section, we present qualitative case studies across a spectrum of time series understanding and reasoning tasks. These examples serve to empirically demonstrate the capabilities of \model\xspace compared to three competitive baselines: Gemini 3 Pro (Visual-centric), ChatTS (Numerical-centric), and GEM (Numerical+Visual).

\begin{figure}[h]
    \centering
    \includegraphics[width=0.75\linewidth]{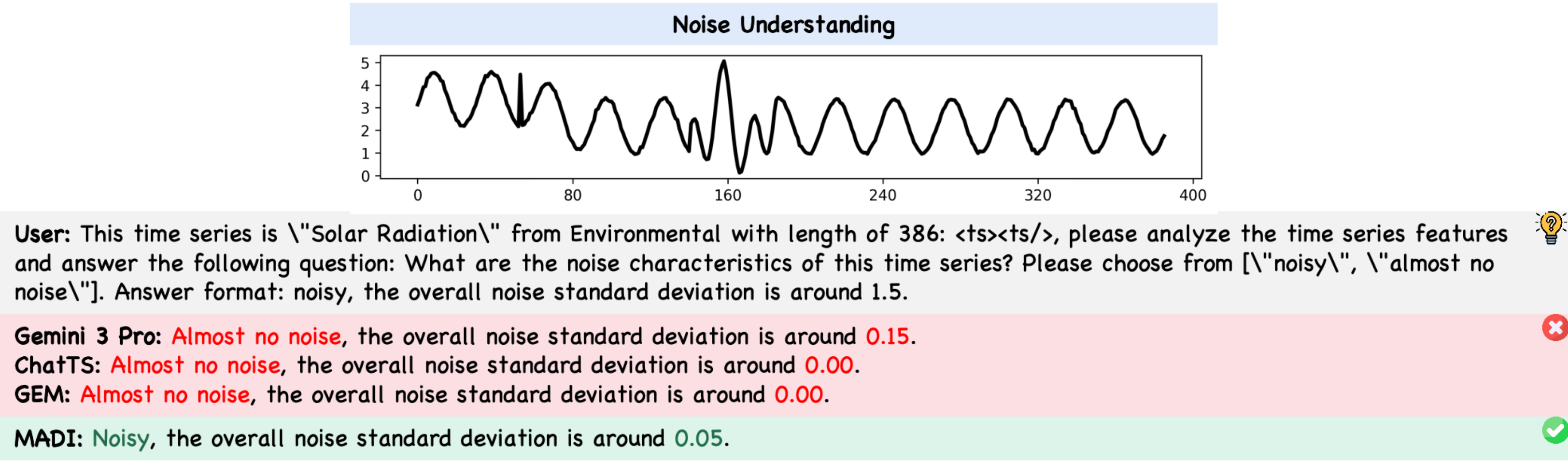}
    \caption{Case study of noise understanding.}
    \label{fig:case_noise}
\end{figure}

\textbf{Noise Understanding.} \figref{fig:case_noise} evaluates noise quantification using solar radiation data, a domain characterized by subtle, high-frequency fluctuations. Although baseline models, such as Gemini 3 Pro, fail to perceive these fine-grained perturbations—incorrectly classifying the series as noiseless, \model\xspace exhibits superior sensitivity. It accurately detects the ``noisy'' nature of the signal and provides a precise standard deviation estimate of 0.05. This result underscores the model's ability to distinguish inherent signal stochasticity from clean trends, a capability often lacking in existing methods.

\begin{figure}[h]
    \centering
    \includegraphics[width=0.75\linewidth]{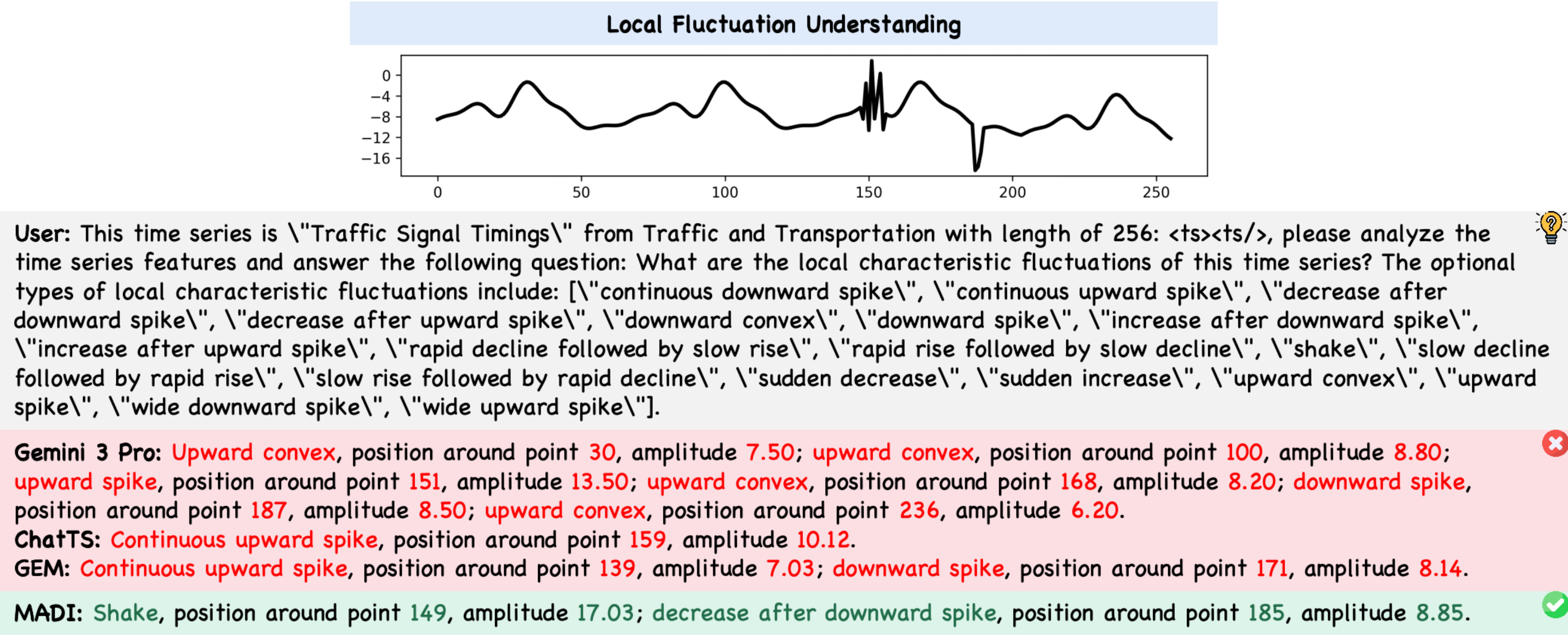}
    \caption{Case study of local fluctuation understanding.}
    \label{fig:case_local}
\end{figure}

\textbf{Local Fluctuation Understanding.} \figref{fig:case_local} illustrates the challenge of characterizing morphological features within traffic signal data. The series exhibits complex local behaviors, specifically a rapid ``shake'' followed by a sharp decline. Baselines like Gemini 3 Pro and ChatTS oversimplify these patterns, hallucinating generic ``convex'' shapes or ``upward spikes'' In contrast, \model\xspace demonstrates high morphological fidelity, accurately localizing and classifying the ``shake'' at point 149 and the ``decrease after downward spike'' at point 185. This highlights the model's precision in parsing fine-grained structural changes in complicated temporal data.

\begin{figure}[h]
    \centering
    \includegraphics[width=0.75\linewidth]{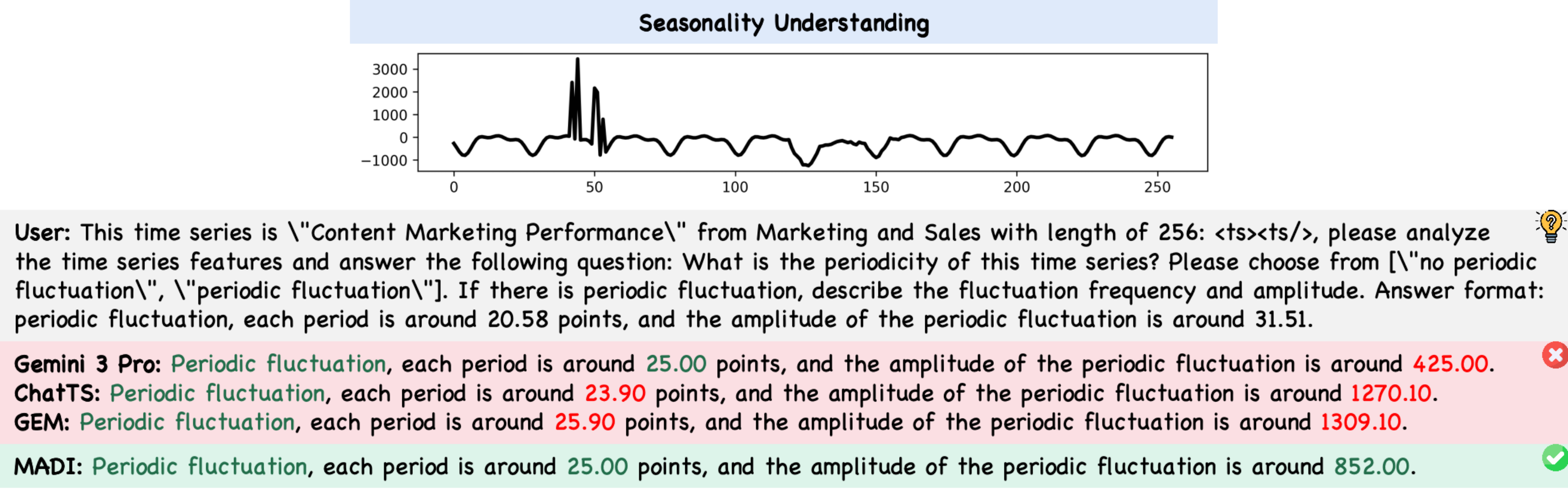}
    \caption{Case study of seasonality understanding.}
    \label{fig:case_season}
\end{figure}

\textbf{Seasonality Understanding.} \figref{fig:case_season} examines seasonality extraction in content marketing performance data, where regular cycles are disrupted by a significant anomaly. This irregularity causes baselines to falter: ChatTS and GEM are sensitive to the outlier, resulting in a gross overestimation of amplitude, while Gemini 3 Pro underestimates it. \model\xspace, however, proves robust to such contamination, correctly identifying the period (25.00) and providing a reliable amplitude estimate (852.00). This demonstrates the model's capacity to identify true periodic features from sporadic irregularities.

\begin{figure}[h]
    \centering
    \includegraphics[width=0.75\linewidth]{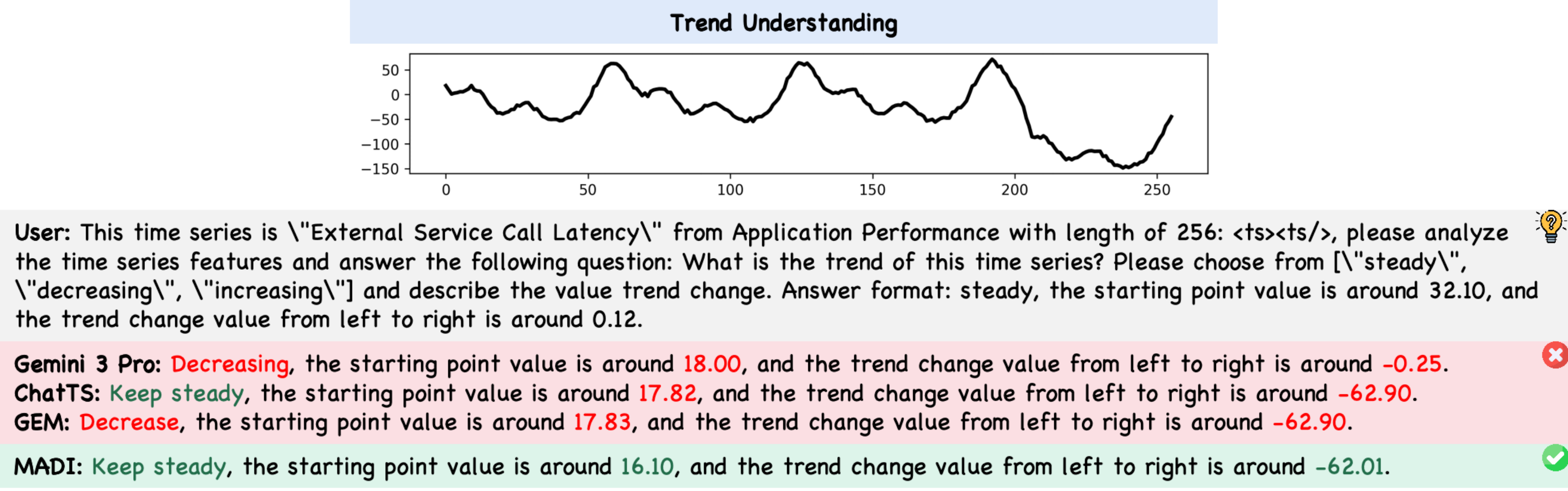}
    \caption{Case study of trend understanding.}
    \label{fig:case_trend}
\end{figure}

\textbf{Trend Understanding.} \figref{fig:case_trend} depicts a trend analysis task on external service call latency data. The series is defined by high volatility and a late-stage downward shift. Although baselines like Gemini 3 Pro and GEM misinterpret the global behavior as strictly ``Decreasing'', and ChatTS fail to capture the precise values of the starting point and global shift, \model\xspace correctly identifies the overall ``Steady'' state while accurately quantifying the specific trend change value (-62.01). This validates the model's ability to maintain a global perspective despite local volatility.

\begin{figure}[h]
    \centering
    \includegraphics[width=0.75\linewidth]{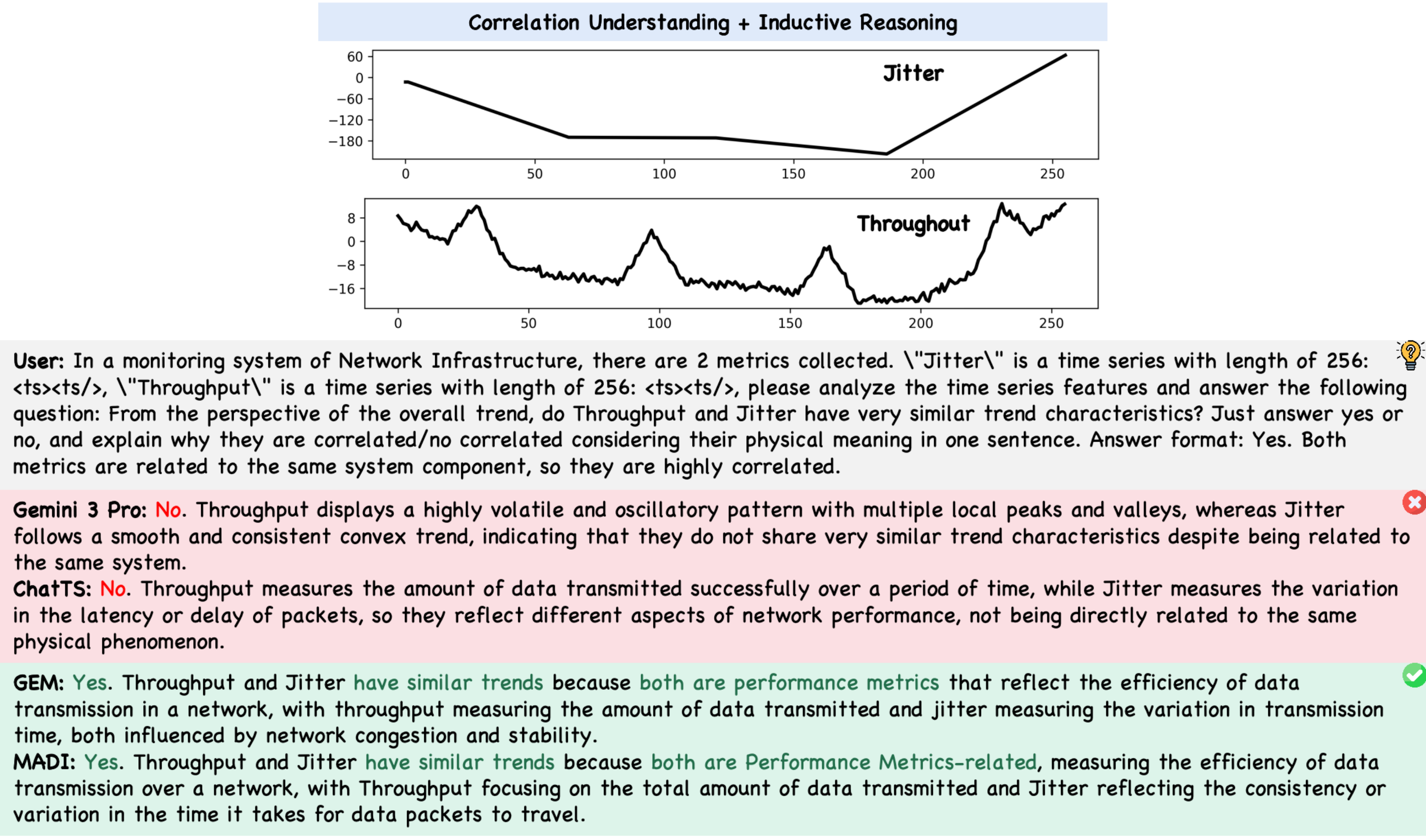}
    \caption{Case study of correlation understanding and inductive reasoning.}
    \label{fig:case_correlation_inductive}
\end{figure}

\textbf{Correlation Understanding and Inductive Reasoning.} \figref{fig:case_correlation_inductive} presents a complex task requiring both correlation analysis and inductive reasoning on network ``Jitter'' and ``Throughput'' metrics. Baselines (\eg Gemini 3 Pro, ChatTS) often rely on superficial signal characteristics (\eg smoothness) or rigid definitions to incorrectly deny a relationship. \model\xspace, conversely, looks past local signal disparities to identify the shared long-term trends driven by the underlying system physics. This confirms the model's ability to perform semantic alignment between coupled time series variables.

\begin{figure}[h]
    \centering
    \includegraphics[width=0.75\linewidth]{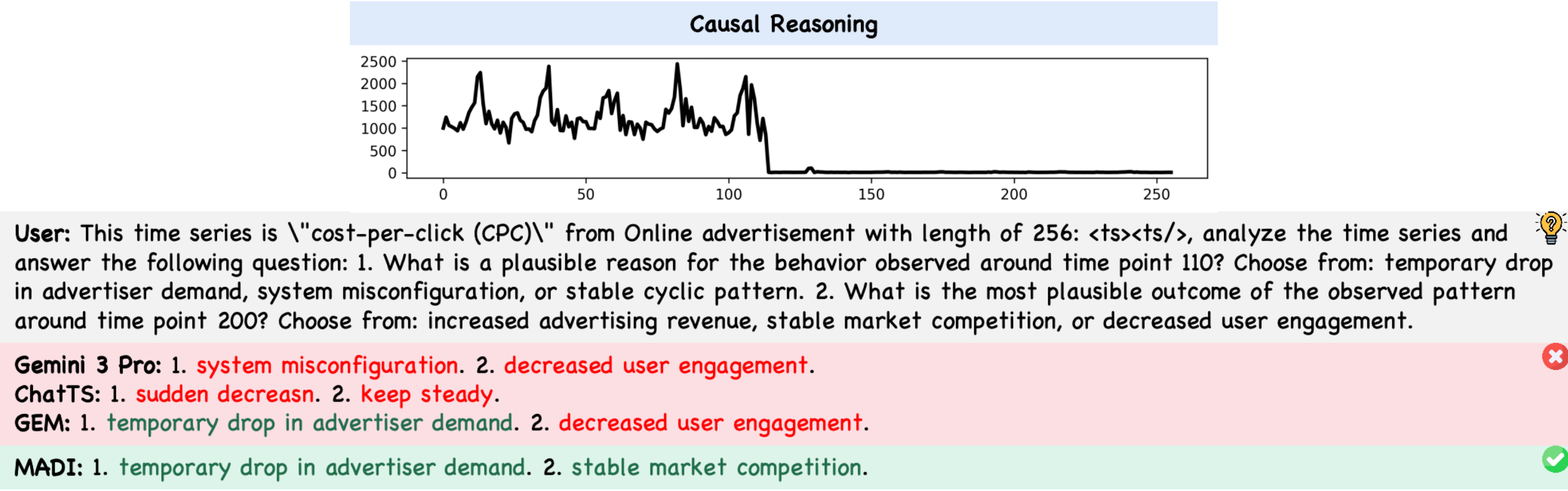}
    \caption{Case study of causal understanding.}
    \label{fig:case_causal}
\end{figure}

\textbf{Causal Reasoning.} \figref{fig:case_causal} demonstrates a causal reasoning scenario using cost-per-click (CPC) data. The series features a precipitous drop around point 110, stabilizing at a lower level. \model\xspace outperforms baselines by grounding its reasoning in domain logic, correctly attributing the shift to a ``temporary drop in advertiser demand'' and predicting a subsequent state of ``stable market competition'' Other models, such as Gemini 3 Pro, hallucinate technical failures (``system misconfiguration''), revealing a lack of domain-aware causal inference.

\begin{figure}[h]
    \centering
    \includegraphics[width=0.75\linewidth]{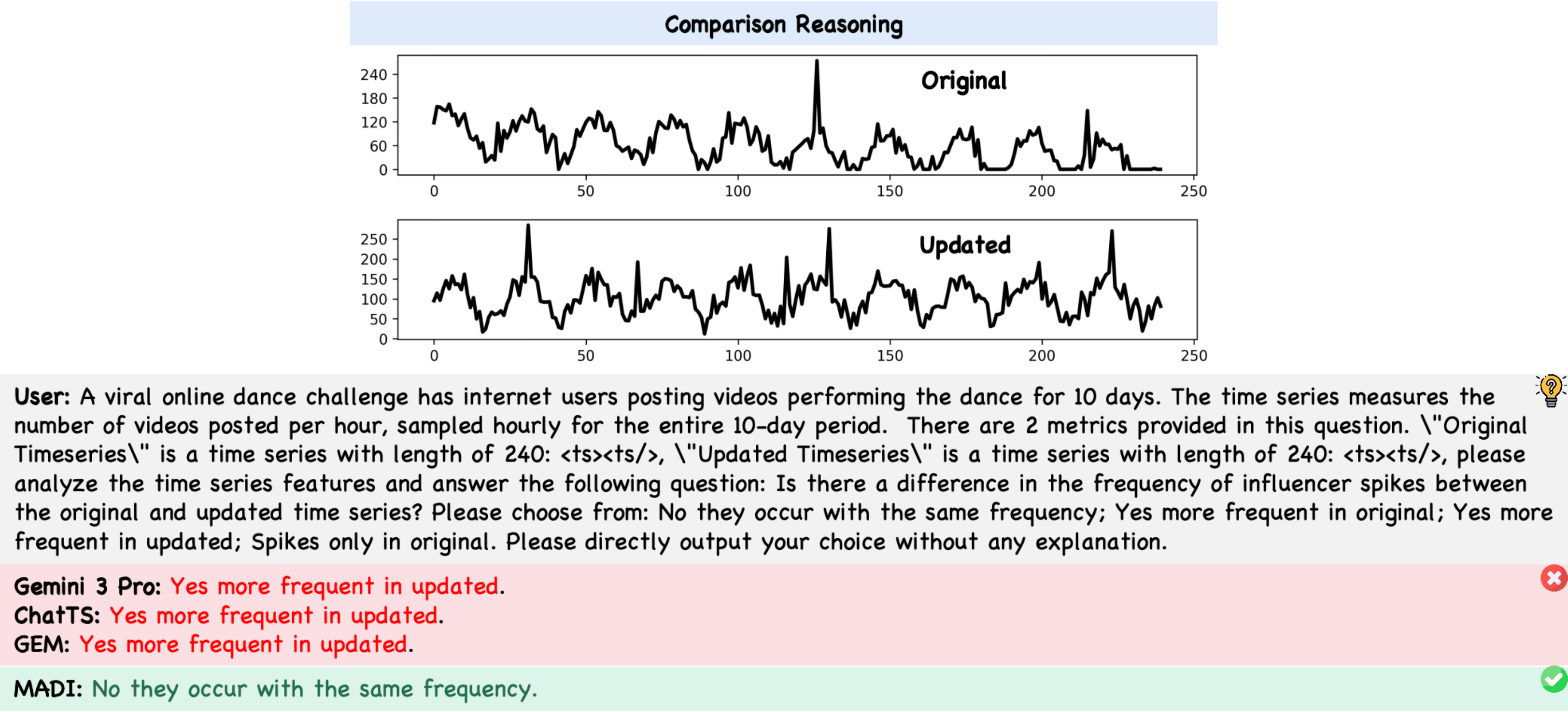}
    \caption{Case study of comparison understanding.}
    \label{fig:case_comparison}
\end{figure}

\textbf{Comparison Reasoning.} \figref{fig:case_comparison} involves a comparative analysis of influencer spike frequency across two video upload time series. A common pitfall for baselines is conflating signal amplitude with event frequency. Consequently, all baselines (Gemini 3 Pro, ChatTS, GEM) hallucinate an increased frequency in the ``Updated'' series due to its visual characteristics. \model\xspace avoids this error, correctly observing that the spikes occur with the same frequency despite amplitude variations, effectively decoupling temporal rate from intensity.

\begin{figure}[h]
    \centering
    \includegraphics[width=0.75\linewidth]{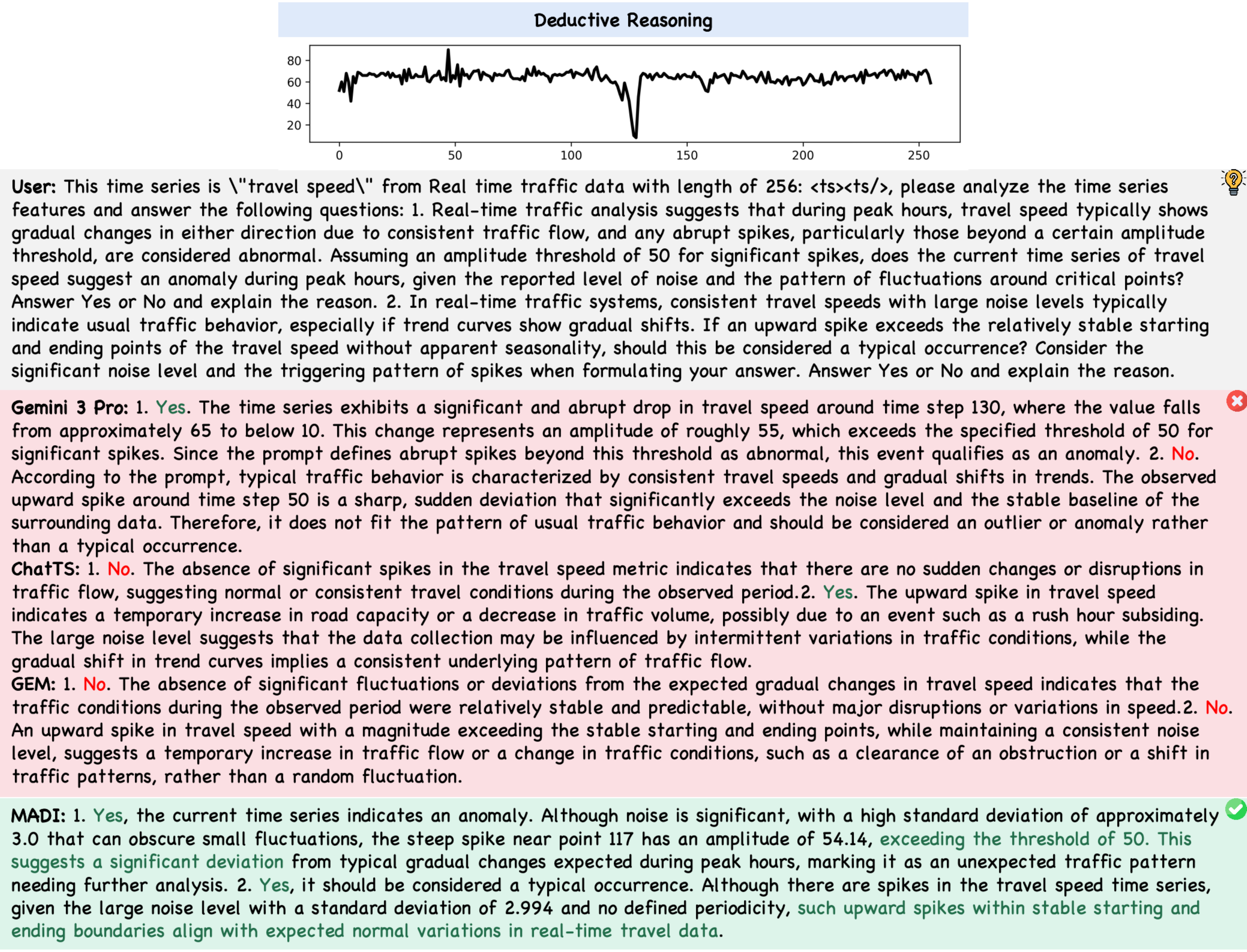}
    \caption{Case study of deductive understanding.}
    \label{fig:case_deductive}
\end{figure}

\textbf{Deductive Reasoning.} \figref{fig:case_deductive} tests the model's ability to apply strict logical constraints to travel speed data, specifically regarding anomaly thresholds and noise patterns. While ChatTS and GEM fail to detect the anomaly, and Gemini 3 Pro fails to logically deduce the ``typical'' nature of the high-noise behavior in the follow-up question, \model\xspace exhibits robust deductive capabilities. It explicitly verifies the amplitude condition (54.14 $>$ 50) to detect the anomaly and correctly reasons that the observed fluctuations align with normal traffic variations given the problem's specific constraints.

\end{document}